\documentclass[journal]{IEEEtran}

\usepackage{cite}
\usepackage{amsmath,amssymb,amsfonts}
\usepackage{algorithmic}
\usepackage{graphicx}
\usepackage[table,xcdraw]{xcolor}
\usepackage{tabularx}
\usepackage{multirow}
\usepackage{amsmath}
\usepackage[english]{babel}
\ifCLASSINFOpdf
\else
\fi
\begin{document}
%
\title{The self-supervised spectral-spatial attention-based transformer network for automated, accurate prediction of crop nitrogen status from UAV imagery}
%
%
%

\author{Xin Zhang, Liangxiu Han*, Tam Sobeih, Lewis Lappin, Mark Lee, Andew Howard and Aron Kisdi
\thanks{Xin Zhang, Liangxiu Han, Tam Sobeih are with the Department of Computing, and Mathematics, Manchester Metropolitan University, Manchester M15GD, U.K (e-mail: x.zhang@mmu.ac.uk; l.han@mmu.ac.uk;T.Sobeih@mmu.ac.uk)}
\thanks{Lewis Lappin and Aron Kisdi is the GMV, Glasgow, Scotland G431QQ, U.K (lewis.lappin@gmvnsl.com;akisdi@gmvnsl.com)}
\thanks{Mark Lee is with the Department of Health Studies, Royal Holloway, University of London, Egham, Surrey, TW200EX, U.K (e-mail:mark.lee@rhul.ac.uk) }
\thanks{Andew Howard is with the Bockhanger Farms Ltd, Oaklands Farm, Ashford TN261ER U.K (e-mail:bockhanger@btconnect.com)}
\thanks {Corresponding author*: L. Han (e-mail: l.han@mmu.ac.uk)}
\thanks{Manuscript received XXX; revised XXX}}

%
%

\markboth{Journal of \LaTeX\ Class Files,~Vol.~13, No.~9, September~2014}%
{Shell \MakeLowercase{\textit{Xin Zhang et al.}}: The self-supervised spectral-spatial attention-based transformer network for automated, accurate prediction of crop nitrogen status from UAV imagery}
%



\maketitle

\begin{abstract}
Nitrogen (N) fertilizer is routinely applied by farmers to increase crop yields. At present, farmers often over-apply N fertilizer in some locations or at certain times because they do not have high-resolution crop N status data. N-use efficiency can be low, with the remaining N lost to the environment, resulting in higher production costs and environmental pollution. Accurate and timely estimation of N status in crops is crucial to improving cropping systems' economic and environmental sustainability. Destructive approaches based on plant tissue analysis are time consuming and impractical over large fields. Recent advances in remote sensing and deep learning have shown promise in addressing the aforementioned challenges in a non-destructive way. In this work, we propose a novel deep learning framework: a self-supervised spectral-spatial attention-based vision transformer (SSVT). The proposed SSVT introduces a Spectral Attention Block (SAB) and a Spatial Interaction Block (SIB), which allows for simultaneous learning of both spatial and spectral features from UAV digital aerial imagery, for accurate N status prediction in wheat fields. Moreover, the proposed framework introduces local-to-global self-supervised learning to help train the model from unlabelled data. The proposed SSVT has been compared with five state-of-the-art models including: ResNet, RegNet, EfficientNet, EfficientNetV2 and the original vision transformer on both testing and independent datasets. The proposed approach achieved high accuracy (0.96) with good generalizability and reproducibility for wheat N status estimation.
\end{abstract}


%
\IEEEpeerreviewmaketitle

\section{Introduction}

\par Nitrogen is an essential plant nutrient and is vital for plant growth and development. The application of N fertilizers has revolutionized farming, increasing crop yields and food production to meet the nutritional needs of billions of people. It is estimated that global nitrogen fertilizer demand was 110 Million Tonnes (MT) in 2015 and is projected to be 120 MT in 2020, costing farmers over \$100 billion per year \cite{FAO2017World,Good2018Toward}. Optimal application of N fertilizers enhances soil fertility and increases crop yields. On the other hand, excessive N inputs are costly for farmers but do not deliver any additional yield benefits, instead resulting in the pollution of natural ecosystems, increases in emissions of the potent greenhouse gas, nitrous oxide, and reductions in biodiversity \cite{Berger2020Crop,Wang2011Excessive}. Wheat crops invariably require fertilizer to grow optimally, are the world’s most commonly consumed cereal grain, and one of the worldwide staple foods. About 35\%-40\% of the global population depend on wheat as their major food crop\cite{Knoema2021Wheat}. Accurate monitoring of the N status in wheat informs farmer decisions on nitrogen fertilizer application rates and timing. It therefore crucial for the economic and environmental sustainability of cropping systems to support a secure food supply chain.  

\par Many crop N estimation methods have been proposed which can be broadly divided into two approaches: destructive and non-destructive. The destructive methods are mostly based on tissue analysis of plant leaves in the laboratory and are time-consuming and costly, which is impractical when collecting the data over large areas \cite{Berger2020Crop,BenitezRamirez2010Monitoring}. In contrast, non-destructive methods perform estimation proximally and remotely of the crop's N status, in a timely fashion and without causing damage to the plants. 
\par Recently, with the development of remote sensing technology, the optical sensors mounted on unmanned aerial vehicles (UAV), airplanes and satellites provide a non-destructive, rapid, and relatively inexpensive crop N estimation method. These sensors, such as RGB sensors and multi to hyperspectral sensors, capture the data remotely to measure the radiation reflected by plants\cite{wang2017non}. The optical sensors can provide rich spectral information in different spectrum regions, including the visible region (380 -- 700nm, VIS), the near infrared region (700 -- 1300 nm, NIR) and the shortwave infrared region (1300 -- 2500 nm, SWIR). The spectral information in these regions are considered to measure the biological (e.g., photosynthetic pigments, chlorophylls) and morphological (leaf area, canopy density) features of the crop, thus deriving the N content and status\cite{Berger2020Crop}. In general, the measurements from these sensors provides a cubic data format containing spatial information in two dimensions (X-Y axis) and abundant spectral information in the third dimension (Z axis). Depending on the dimensions of the data used, we can classify the estimation methods into two categories: spectral analysis and spatial analysis.

\par The spectral analysis methods in remote sensing applications assume that the spectral information of each pixel can be used to measure the objectives, such as the N content of the crop\cite{imani2020overview}.  One of the most commonly used spectral analysis methods is to use vegetation indexes (VI) based on specific wavelengths to predict the crop N content and status, such as Normalized Difference Vegetation Index (NDVI)\cite{johnson2014assessment} and Leaf area index (LAI)\cite{pierce1994regional}). However, the N content of crops is considered a complex problem, which is affected by various influential factors such as ambient lighting conditions and variations amongst the types of crops. The VI methods based on specific wavelengths, are considered sensitive and lack generalizability \cite{Chlingaryan2018Machine}. Machine learning has shown the effectiveness of solving nonlinear problems from multiple sources \cite{Jordan2015Machine} and in recent years, has been increasingly used for crop N estimation. In \cite{Shi2021Rice} and \cite{Qiu2021Estimation}, the author used three machine learning algorithms (Random Forest (RF), Support Vector Machine (SVM) and Artificial Neural Networks (ANNs)) to estimate rice nitrogen based on all available spectral information, with the random forest (RF) demonstrating high accuracy and strong generalization performance. 

\par With advancements in remote sensing, the resolution of the data, including the spatial and spectral, have been significantly improved. The variation in spectral features between neighbouring pixels increases as the spatial resolution is improved\cite{zhang2020well}. The spatial information in the finer spatial resolution data can be used to measure the structure and health condition of the crop; these are considered essential attributes for characterizing the N status \cite{Roth2018Predicting}. However, conventional spectral-based VI methods have difficulty analysing high spatial resolution data, with few works using spatial information for crop N estimation\cite{Roth2018Predicting}. Accurate estimating of crop N content incorporating spatial information remains a challenge.

\par Over the past few years, with the emergence of graphic processing units (GPU) \cite{Alom2018Improved}, deep learning (DL) methods, have dominated computer vision tasks and are considered superior to pre-existing methods for extracting spatial features from images\cite{Nanni2017Handcrafted}. They are rapidly being used for crop N estimation. In the work \cite{Lewis2020Classification}, the authors proposed a DL classification model for N status prediction in coffee plants. Sethy et al. \cite{Sethy2020Nitrogen} used six leading DL architectures to predict nitrogen deficiency on rice crops. In the work \cite{Tran2019Comparative}, the DL method has been used to classify and predict early N deficiencies during the growth of the tomato plant.

\par However, directly using DL methods to estimate crop N still suffers from the following problems. Firstly, most existing DL structures are designed to capture spatial information with no specific module for spectral information learning, which is important for crop N status estimation. Secondly, the DL models are data-hungry in nature, which require large datasets (labeled data) for model training to achieve good performance and avoid over-fitting. Finally, DL algorithms have a high computational complexity, which do not scale well with remote sensing products taht are usually of a large size.
 
\par In this work, to overcome the aforementioned issues, we propose a self-supervised spectral-spatial attention-based transformer network (SSVT) for automatic and accurate crop N status estimation. Our network is inspired by the state-of-the-art vision transformer (ViT) structure\cite{Wolf2020Transformers:}, which allows for capturing the local to long-range spatial information from images. To the best of our knowledge, this is the first work that explores the transformer network combined with self-supervised learning for accurate crop N status prediction. Our contributions include: 

\begin{enumerate}

\item A novel spectral-spatial attention-based vision transformer is proposed, in which both the spectral and spatial information are considered. A Spectral Attention Block (SAB) is proposed to learn spectral-wise features such as color information. Meanwhile, a Spatial Interact (SIB) is introduced after SBA to learn corresponding spatial information.

\item A Local-to-global self-supervised learning (SSL) method is proposed to pretrain the model on the unlabeled images to resolve the data-hungry paradigm in DL model training and improve the model's generalization performance on independent data.

\item A linear computational complexity is achieved using the cross-covariance matrix instead of the original gram matrix operation in the attention block. It changes the complexity of the transformer layer from quadratic to linear, which makes it possible for the model to handle large size images.
\end{enumerate}
The rest of this paper is organized as follows: Section 2 presents the related work; Section 3 details the proposed method; In Section 4, the experimental evaluation is described; Section 5 concludes the proposed work and highlights the future work.

\section{Related work}

\subsection{Non-destructive Crop N Estimation methods}

\par Over the past two decades, remote sensing technology has been considered one of the most promising methods to provide a non-destructive way in which to measure and estimate crop N content and status in fields and wider environments\cite{scharf2002remote}. The principle behind the technology is that by using optical sensors (e.g. RGB, multi to hyperspectral sensors) mounted on UAVs, airplanes and satellites, accurate information about the morphological and physiological condition of the crops can be measured, which are considered to be related to crop N content \cite{imani2020overview}. These sensor measurements can provide rich spectral information in different spectrum regions, including the visible region (380 -- 700nm, VIS), the red edge region (690 -- 730nm), the near infrared region(700 -- 1300 nm, NIR) and the shortwave infrared region (1300 -- 2500 nm, SWIR). The spectral information in these regions are considered to measure the biological (e.g. photo- synthetic pigments, chlorophylls) and morphological (leaf area, canopy density) features of the crop and thus derive the N content and status\cite{Berger2020Crop}. 

\begin{figure}[h]
    \centering
    \includegraphics[width=0.45\textwidth]{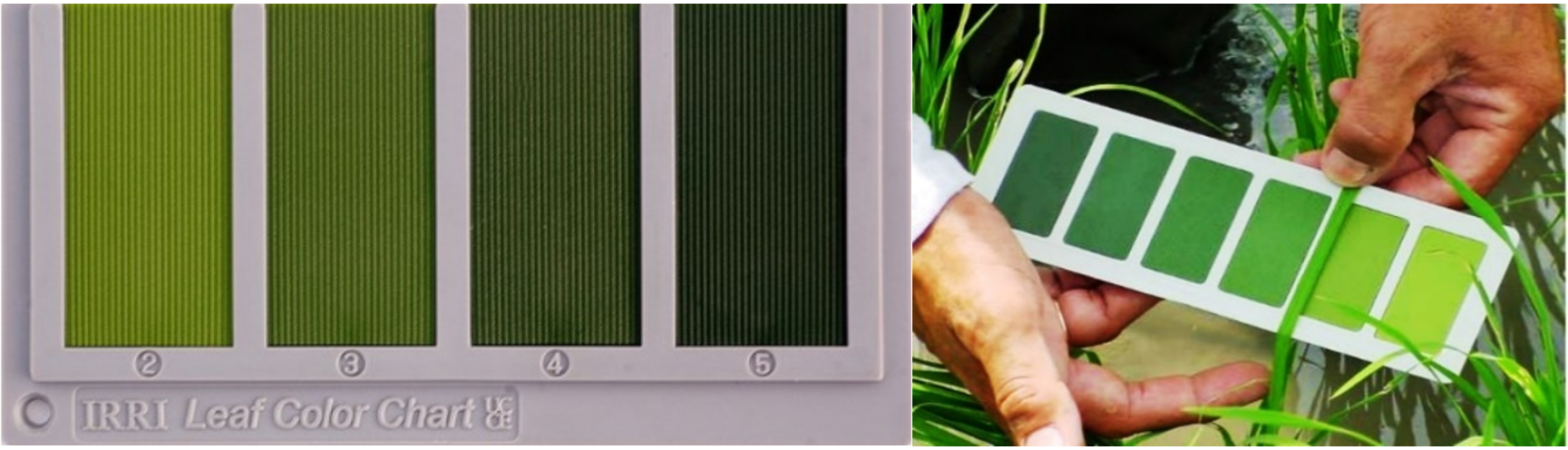}
    \caption{Leaf Color Chart determines the Nitrogen status of paddy rice \cite{Yang2003Using}.}
    \label{FIG:1}
\end{figure}

For instance, the measurements from RGB sensors provide the spectral/color information in visible regions including red, green, and blue wavelengths. They have been used to measure the crop physiological features such as leaf chlorophyll, carotenoids and anthocyanins content, which are closely related to leaf nitrogen content \cite{hunt2013visible,solovchenko2008screening}. The leaf color chart (LCC) is an early stage and commonly used method to determine the N status of crops by using the color information\cite{Yang2003Using}. The LCC has five categories, ranging in color from yellow to green. It determines the nitrogen content of crops based on the degree of green color of rice leaves ({Fig.~\ref{FIG:1}}). The multi to hyperspectral sensor measurements provide a broader range of spectral information, including the red edge , NIR, and SWIR, which have been used to measure not only the biological features, such as the absorption features of proteins, but also the morphological features like the area and density of the leaf and canopy \cite{baret2007lai,hank2019spaceborne}.

\par Generally, the remote sensing imagery captured by the optical sensors provides a cubic data format containing spatial information in two dimensions (X-Y axis) and abundant spectral information in the third dimension (Z axis). Depending on the dimensions of the data used, we can classify the estimation methods into two categories: \textbf{spectral analysis and spatial analysis}.

The spectral analysis approaches are mainly based on the spectral information of each pixel to distinguish, identify or measure objectives. 
So far, based on the abundant spectral information, many studies have been developed to estimate crop N content and status. The most widely used methods are the vegetation index (VI)-based methods focusing on specific bands. These bands are chosen to estimate N status based on their sensitivity to the chlorophyll content, leaf area and canopy density, such as the green wavelength (550nm), red wavelength (675nm), red edge wavelength (720nm) and NIR wavelength (905nm) \cite{padilla2018proximal,clevers2013remote,afandi2016nitrogen}. Once validated, these methods produce linear indicator indices from the selected bands to measure the N status of the crops. In the work \cite{Pagola2009New}, a greenness index (GI) using RGB wavelength of the colour image was proposed to estimate the amount of N in the plant. In \cite{johnson2014assessment}, the Normalized Difference Vegetation Index (NDVI) was used to estimate N status of corn and soybean in the United States. Glenn et al.\cite{fitzgerald2010measuring} introduced a canopy chlorophyll content index (CCCI) to measure and predict canopy nitrogen in wheat. However, the VI-based methods only utilized specific bands relevant to crop N, the rest of the spectral information was not exploited, especially for the multi to hyper spectral sensor measurements where a mass of information was ignored. These types of methods are sensitive to the crop types and the growing stages, and lack generalizability \cite{Chlingaryan2018Machine}. During the last decade, machine learning (ML) approaches have shown the effectiveness of solving complicated, nonlinear problems from multiple sources \cite{Jordan2015Machine} and have been increasingly used for crop N estimation in recent years. In \cite{wang2017non}, several ML algorithms such as Principal Components Regression (PCR), Partial Least Squares Regression (PLSR) and Stepwise Multiple Linear Regression (SMLR) were used to extract useful features to estimate leaf N content from from all the available wavelengths simultaneously. In the research \cite{Shi2021Rice}, simple nonlinear regression (SNR), backpropagation neural network (BPNN), and random forest (RF) regression were used to determine the rice N nutrition status with RGB images. In works \cite{Qiu2021Estimation} and \cite{Zha2020Improving}, the authors used support vector machine (SVM), multiple linear regression (SMLR), and Artificial Neural Networks (ANNs) to estimate rice nitrogen nutrition index with UAV RGB images. The work \cite{Mehra2016predicting} used ANNs and RF to predict the biotic stress of winter wheat. A review research \cite{Chlingaryan2018Machine} indicated that ML approaches would result in more cost-effective and comprehensive solutions for a better crop N status assessment.

\par However, with the development of remote sensing technologies, the spatial resolution of the data has been significantly improved. As the spatial resolution of the data increases, the consistency of the spectral information between pixels decreases, leading to a reduction in the performance of the conventional spectral analysis methods\cite{zhang2020well}. Moreover, the spatial information in the finer spatial resolution data can be used to measure the structure and health condition of the crop; these are considered essential attributes for characterizing the N status \cite{Roth2018Predicting}. Currently, only a small amount of hand-crafted spatial information, such as canopy cover, are used in N status estimation \cite{Lee2011Estimating,Li2010Estimating,Roth2018Predicting,Zhao2021Estimating}.  Therefore, accurately estimating crop N content incorporating spatial information remains a challenge.

\par Over the past few years, convolutional neural networks (CNN) have dominated computer vision tasks \cite{Alom2018Improved}. Unlike standard hand-crafted feature learning methods, the CNN, as a filter bank, can automatically extract spatial features from a local receptive field in images \cite{Nanni2017Handcrafted}. Azimi \cite{Azimi2021deep} proposed a 23-layered CNN to measure the crop stress level in plants due to nitrogen deficiency and found that CNNs outperform most machine learning methods in fast \& accurate identification of stress in plants. Lee \cite{Lee2017How} proposed a hybrid global-local feature extraction model to extract spatial features of the leaves to perform plant classification. Their results showed the strength of detecting spatial features using CNNs as compared to hand-crafted features. Meanwhile, it was found that traditional CNNs could only extract local spatial information \cite{Islam2020How} and failed to capture long-range global spatial information. Therefore, a new analysis method that can capture both spectral and spatial information from remote sensing imagery for crop N estimation is important. 

\subsection{Vision Transformer}
\par Recently, Vision Transformer (ViT) \cite{Wolf2020Transformers:} has attracted increasing attention in computer vision tasks due to its capability to capture long-range spatial interactions as well as introducing less inductive bias, compared to widely-used convolutional neural networks (CNNs). It has been considered to be a solid alternative for CNNs. The essence of ViT is to use a self-attention scheme \cite{Vaswani2017Attention} to capture long-range dependencies or global information, focusing on spatial information. 
\begin{figure}[h]
    \centering
    \includegraphics[width=0.4\textwidth]{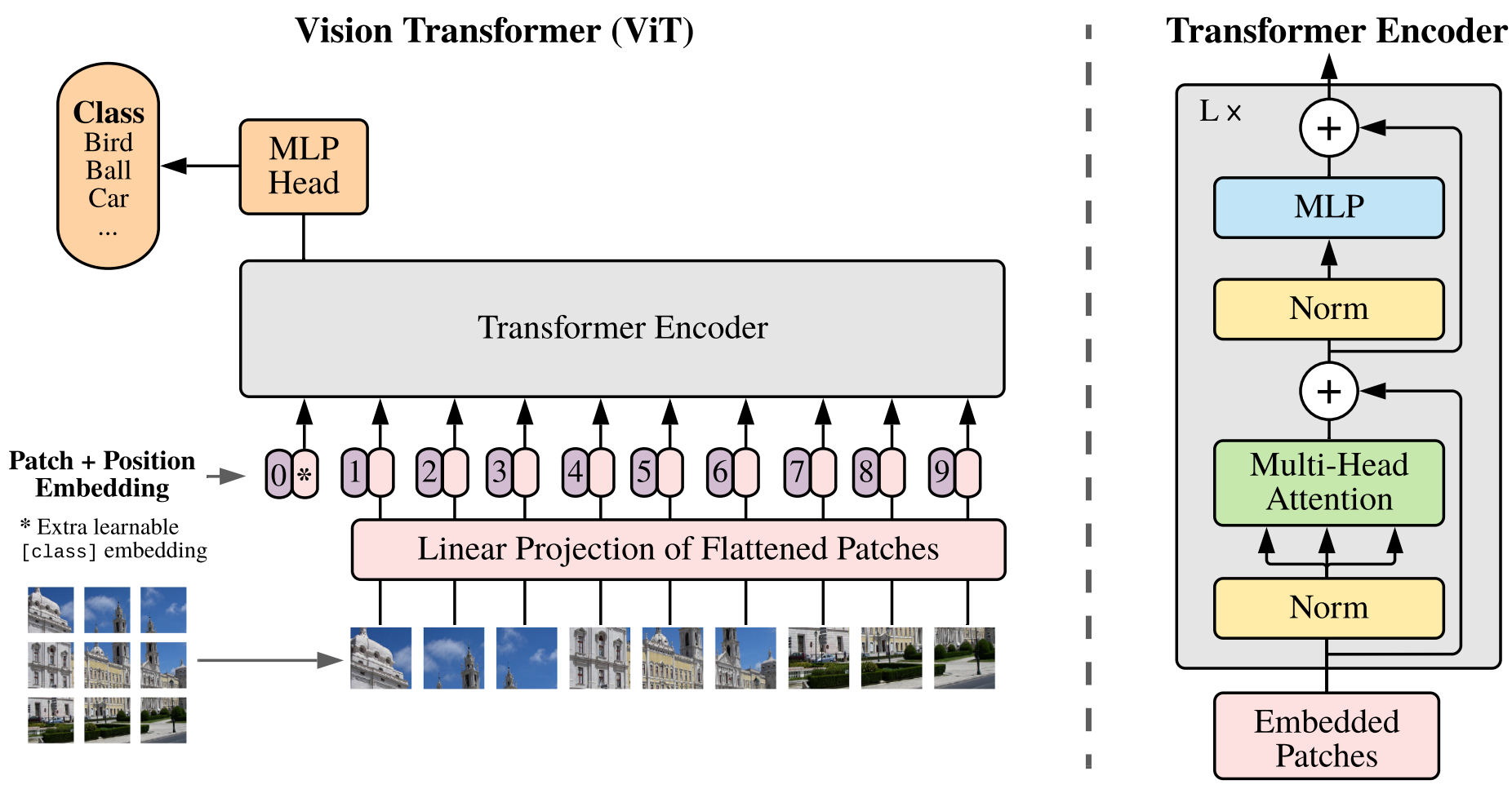}
    \caption{The structure of Vision Transformer \cite{Dosovitskiy2020Image}.}
    \label{FIG:2}
\end{figure}
\par There are four main parts in the transformer encoder as shown in {Fig.~\ref{FIG:2}}: Multi-Head Self Attention Layer (MSP), Multi-Layer Perceptrons (MLP), Layer Norm, and Residual connections introduced in CNN evolution. The MSP is the core of the transformer. It allows the model to integrate information globally across the entire image. It is used to concatenate the multiple attention outputs linearly to expected dimensions. The multiple attention heads help learning local and global dependencies in the image. MLP contains two fully connected layers with Gaussian Error Linear Unit (GELU) as an essential part of the transformer that stops and drastically slows down rank collapse in model training \cite{Dong2021Attention}. Layer Norm is the normalization method in the NLP area instead of Batchnorm in vision tasks. It is applied before every block as it does not introduce any new dependencies between the training images. This helps to improve training time and generalization performance. Residual connections are applied after every block as they allow the gradients to flow through the network directly without passing through nonlinear activations.

However, the ViT network cannot be used to estimate Crop N status directly. The ViT has the ability to extract spatial information of an image, but it can not extract spectral information, which has been proven to contain the most important features related to the crop N status. Moreover, the ViT has a quadratic computational complexity to the image size, which limits its application on large images and requires large-scale training datasets (i.e., JFT-300M) to perform well\cite{Touvron2021Training}. The SSL technology, which allows models to be trained with unlabelled data, is considered to solve this latter problem\cite{liu2021self}. 

\subsection{Self-supervised learning (SSL)}
Acquiring extensive, labeled data for training DL models is challenging. Self-supervised learning provides an effective way to enable learning from large amounts of unlabeled data. SSL can be broadly divided into Generalized Adversarial Networks (GANs) and Contrastive learning \cite{Liu2021Self-supervised}. GANs are unsupervised learning tasks that involve automatically discovering and learning the regularities or patterns in input data in such a way that the model can be used to generate new samples \cite{Goodfellow2014Generative}. Unlike generative models, Contrastive learning \cite{Hadsell2006Dimensionality} is used to determine which representations attract comparable samples and which ones repel them. The representations from contrastive self-supervised pretraining can be used in specific supervised downstream vision tasks. 
Generally, contrastive SSL usually consists of three parts: 1) image augmentation, 2) feature extraction/encoder, and 3) contrastive loss to quantify the similarity between representations. Image augmentation creates positive pairs by generating different augmented views of the same image, such as color augmentation, image rotation/cropping and other geometric transformations. 
Then a CNN network is used to encode the augmented images as vector representations. The Siamese Neural Network \cite{Bromley1993Signature} is the most widely used neural network architecture to find the similarity between the representations in contrastive learning. It contains two or more identical subnetworks. Each sub-network has the same architecture with the same parameters and weights. Parameter updating is mirrored across both sub-networks. In general, training in Siamese Neural Network is compared against a positive pair and a negative pair. The negative vector pair is used for learning in the network, while the positive pair acts in a regularization role. The negative pairs rely on different images, which are hard to define.
An evolutional work (BYOL) retains the Siamese architectures but eliminates the requirement of negative samples \cite{Grill2020Bootstrap}. BYOL proposed a momentum training that rolling weight updates as a way to give contrastive signals to the training. Recent methods such as SwAV \cite{Caron2021Unsupervised}, MoCo \cite{He2020Momentum}, and SimCLR \cite{Chen2020Simple} with modified configurations have produced results comparable to the state-of-the-art supervised method on the ImageNet public dataset. 
However, most SSL methods are mainly based on standard convolutional networks. The SSL for vision transformer models are new. In this work, inspired by BYOL, we proposed a local-to-global SSL for the vision transformer network.

\section{Method and Materials}

\subsection{Dataset description}
\begin{figure}[h]
    \centering
    \includegraphics[width=0.4\textwidth]{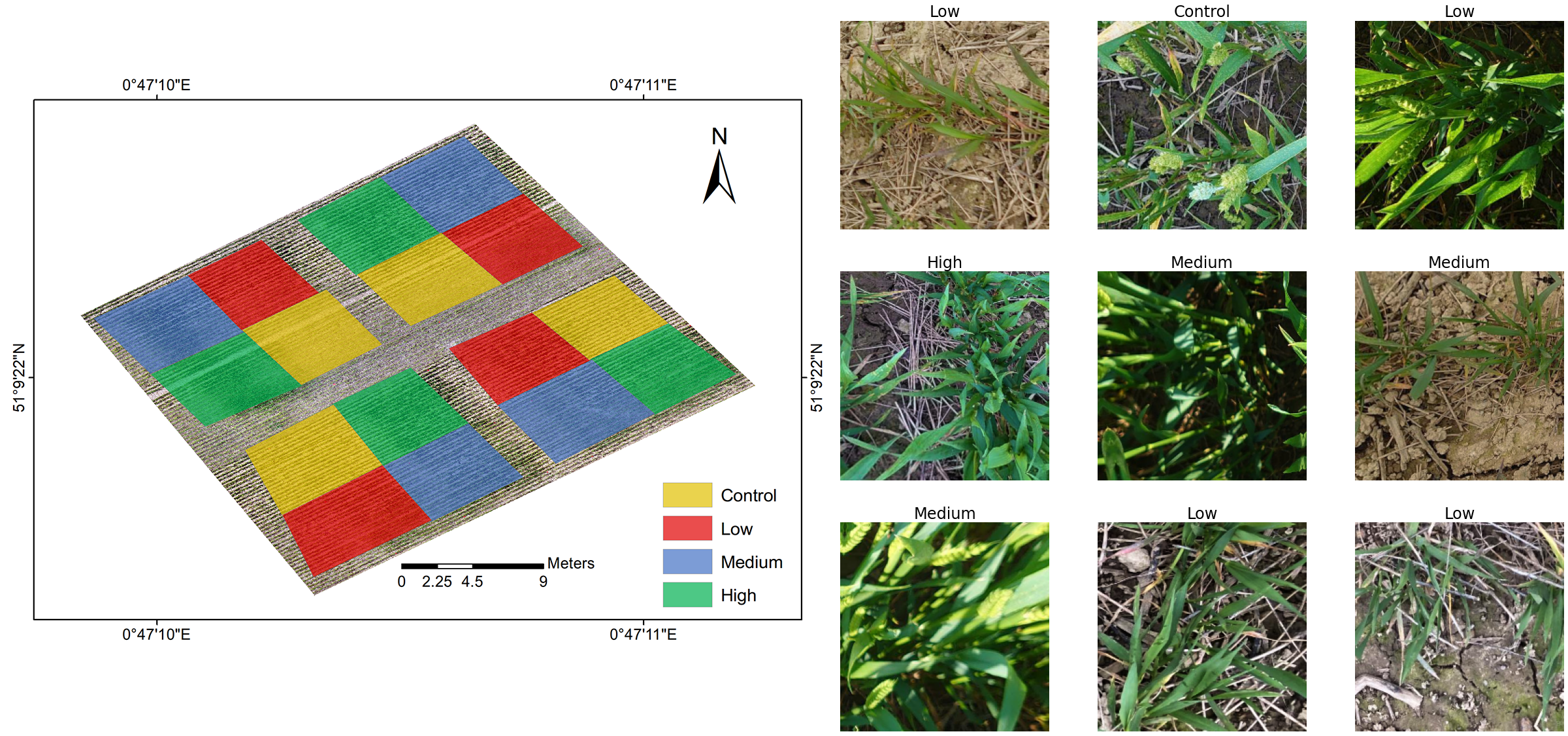}
    \caption{a) Experimental layout, with plots randomly allocated into four treatments, split into four blocks. Treatments were high (green), medium (blue) and low fertilizer rates (red), and an unfertilized control (yellow). b) Images collected from different plots with different treatments.}
    \label{FIG:5}
\end{figure}

In this work, we have collected the data at a controlled wheat field located near Ashford, south-eastern UK (51.156N, 0.876E) ({Fig.~\ref{FIG:5}}). We adopted a 4 x 4 factorial design in the controlled field experiment, with four randomly allocated N treatments replicated within four blocks, totaling 16 plots of 16 $m^2$ (4m x 4m). The plots were established prior to the first fertilizer application. The four treatments were low (80 $kg N ha^-1 yr^-1$), medium (160 $kg N ha^-1 yr^-1$), and high (240 $kg N ha^-1 yr^-1$) fertilizer rates, with unfertilized control. These values were chosen because they were representative of application rates commonly used by arable farmers. Five applications were used to add N fertilizer to the plots every three weeks between February – June 2021.

\begin{table}[h]
\caption{Monitoring schedule and Data collection summary}\label{table:1}
\centering
\resizebox{0.45\textwidth}{!}{%
\begin{tabular}{cccc}
\textbf{Growing Stage}                                                         & \textbf{Date}                  & \textbf{Field image collection} & \textbf{Drone}            \\
\rowcolor[HTML]{F2F2F2} 
\cellcolor[HTML]{F2F2F2}                                                       & 21-Mar                         & 160                             & $\checkmark$                          \\
\cellcolor[HTML]{F2F2F2}                                                       & 02-Apr                         & 163                             & $\checkmark$                          \\
\rowcolor[HTML]{F2F2F2} 
\multirow{-3}{*}{\cellcolor[HTML]{F2F2F2}\textbf{Tillering \& Stem Extension}} & 09-Apr                         & 177                             & $\checkmark$                          \\
                                                                               & 06-May                         & 166                             & $\checkmark$                         \\
                                                                               & \cellcolor[HTML]{F2F2F2}14-May & \cellcolor[HTML]{F2F2F2}177     & \cellcolor[HTML]{F2F2F2}$\checkmark$ \\
\multirow{-3}{*}{\textbf{Heading \& Flowering}}                                & 24-May                         & 175                             & $\checkmark$                          \\
\rowcolor[HTML]{F2F2F2} 
\cellcolor[HTML]{F2F2F2}                                                       & 07-Jun                         & 177                             & $\checkmark$                          \\
\cellcolor[HTML]{F2F2F2}                                                       & 22-Jun                         & 177                             & $\checkmark$                          \\
\rowcolor[HTML]{F2F2F2} 
\multirow{-3}{*}{\cellcolor[HTML]{F2F2F2}\textbf{Ripening \& Maturity}}        & 28-Jun                         & 175                             & $\checkmark$                         
\end{tabular}
}
\end{table}

\begin{figure}[h]
    \centering
    \includegraphics[width=0.45\textwidth]{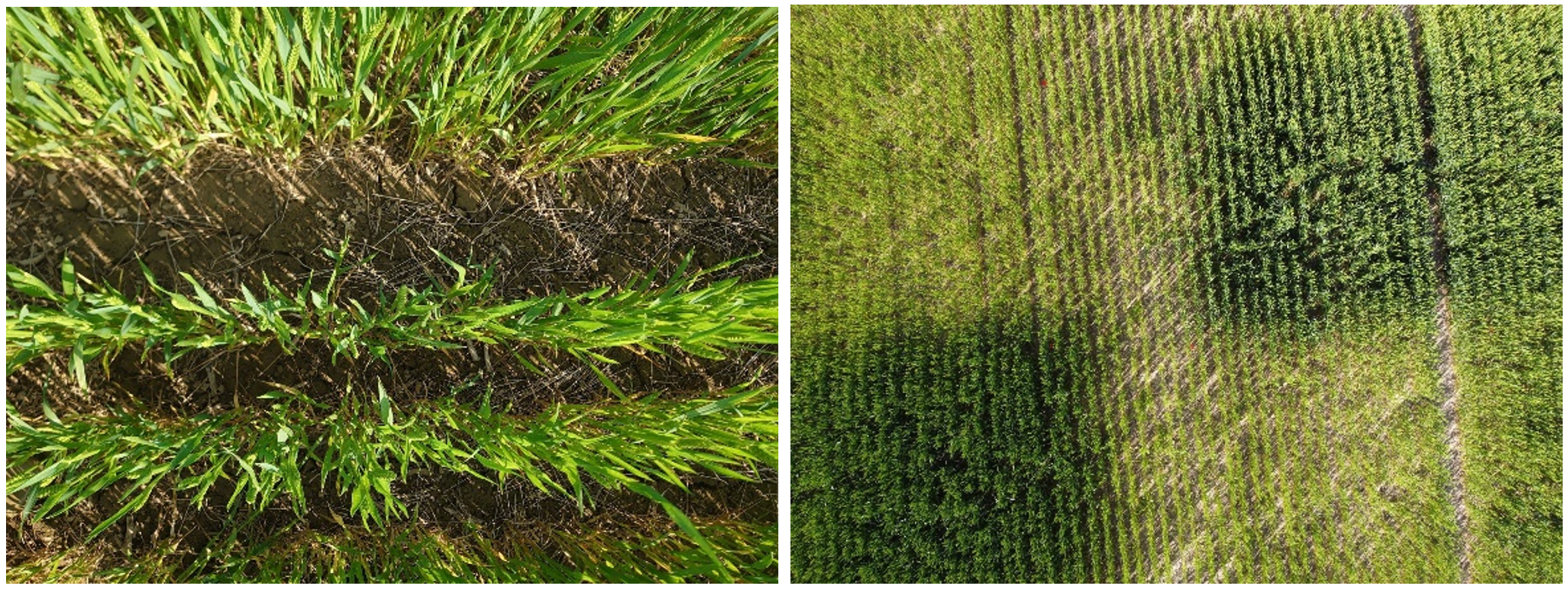}
    \caption{The collected images. The left one is collected near the ground. The right one is collected from the drone.}
    \label{FIG:6}
\end{figure}

Two types of digital camera images were collected at the canopy scale, via near-ground sensing and UAV-based remote sensing, from these plots during all the wheat growing stages, including Tillering \& Stem Extension, Heading \& Flowering and Ripening \& Maturity. A Sony Xperia 5 with a 12-megapixel Exmor RS CMOS is used to collect the near-ground images with a focal length of 26mm. A DJI MAVIC pro with 12.35-megapixel CMOS is selected to capture the images from the air with a focal length of 26mm.
The detailed monitoring schedule is shown in {Table.~\ref{table:1}}. {Fig.~\ref{FIG:6}} shows the sample images. The UAV flight heights were from 10m to 30m. A total of 1449 field images are used in this work. The image size is 4032 x 3024.

\subsection{Method}
In this work, we have proposed a deep learning based framework to accurately estimate the nitrogen status of wheat from remote sensing datasets. This framework consists of two main parts: the spectral-spatial attention vision transformer (SSVT) and a local-to-global self supervised learning method. 

\subsubsection{Spectral-Spatial attention Vision Transformer (SSVT)}

\begin{figure}[h]
    \centering
    \includegraphics[width=0.65\textwidth]{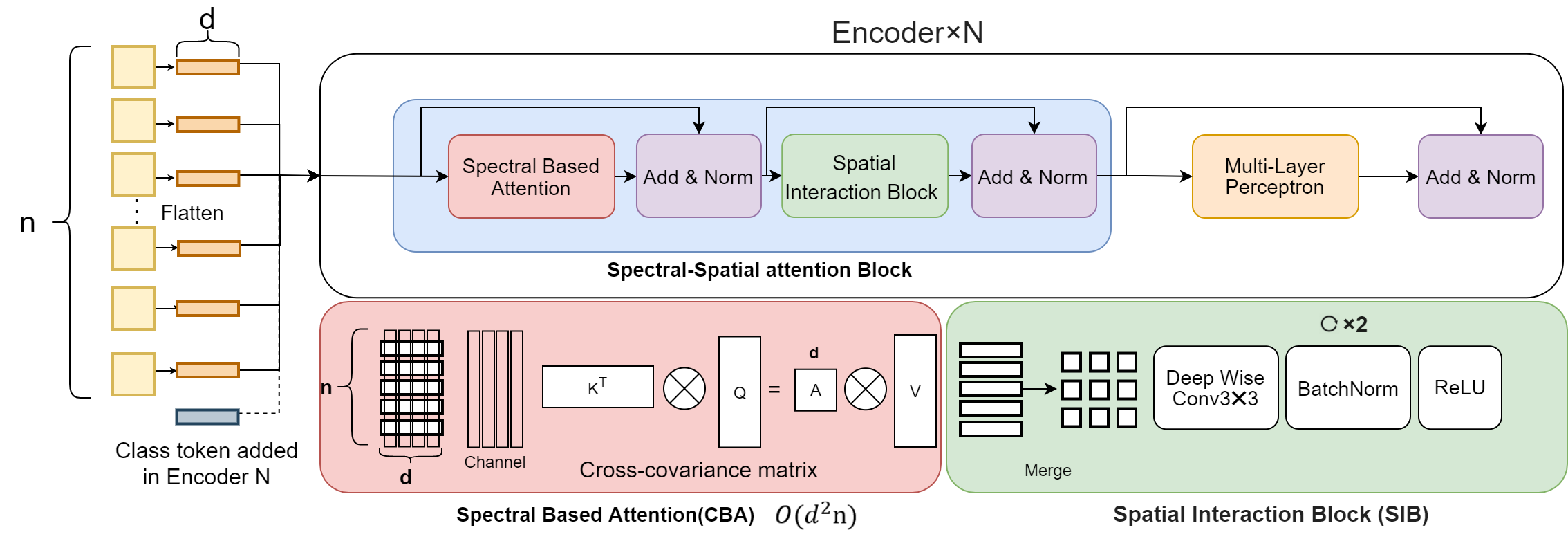}    
    \caption{The structure of the proposed spectral-spatial attention vision transformer (SSVT)}
    \label{FIG:3}
\end{figure}
\par A transformer network named SSVT is developed to accurately estimate the nitrogen status of wheat, capable of capturing both spatial and spectral features from large UAV-based digital aerial imagery. The proposed conceptual architecture is shown in {Fig.~\ref{FIG:3}}. The design rationale is three-fold:
\begin{enumerate}
    \item As shown in previous research \cite{Chlingaryan2018Machine}, spectral information plays a vital role in determining nitrogen status at leaf and canopy scales. In this work, the spectral-based attention block is proposed to learn spectral-wise features such as color information.  
    \item To learn the spatial information, a spatial interaction block is introduced after the spectral-based attention block.
    \item To address the quadratic computing complexity of the ViT, the covariance matrix is used to replace the gram matrix, which can help reduce computational complexity from the quadratic complexity ($O(n2)$) to linear complexity ($O(n)$) where n represents the number of input patches.
\end{enumerate}
The input images are first split into patches and flattened into vectors using a linear projection operation. Then, each vector is regarded as a sequence and fed into the transformer encoders. A class token is added to represent an entire image that can be used for classification. It is actually a vector that is learned during gradient descent. In this work, we add the class token in the last encoder block, which only lets encoders' attention mechanism perform between images. Each encoder consists of two core components, including 1) Spectral-Spatial attention Block, which consists of spectral-based and spatial interaction blocks, and 2) Multi-Layer Perceptron (MLP). 
\paragraph{1) Spectral-Spatial attention Block}
\par In this work, to address the spectral and spatial information in the transformer encoder, we proposed a Spectral-Spatial attention Block (this is different from the Multi-Head Self Attention Layer in the original vision transformer network). The Spectral-Spatial attention Block consists of two main parts: Spectral Based attention (SBA) and Spatial Interaction block (SIB).

\par Spectral Based attention (SBA)
\par The SBA module adopts an attention mechanism with Query, Key and Value. In this work, to decrease the computational complexity of the scaled dot-product attention used by Transformer, cross-covariance is proposed to replace the matrix operation in the self-attention function. Given the packed matrix representations of queries $\ Q\in R^{n\times d}$, keys $K\in R^{n\times d}$ , and values $V\in R^{n\times d}$, the cross-covariance attention is given by:
\begin{small}
\begin{equation}
\mathrm{SBA\ }\left(Q,K,V\right)=A_{XC}\left(K,Q\right)V=SoftMax{\left(\frac{{\hat{K}}^\top\hat{Q}}{\tau}\right)}V
\end{equation}
\end{small}
Where $n$ denotes the number of patches, $d$ denotes the dimensions of keys (or queries) and values, which means the number of pixels in each patch. $A_{XC}\left(K,Q\right)$ denotes as attention matrix, $SoftMax$ is applied in a row-wise manner. Where the attention weights $A_{XC}$ are calculated using a cross-covariance matrix. ${\hat{K}}^\top\hat{Q}$ is the cross-covariance matrix size of $d\times d$. In \cite{El-Nouby2021XCiT:}, the author found that controlling the data range in attention strongly enhances the stability of training, here $\hat{Q}$ and $\hat{K}$ denoted the normalized matrices Q and K. The inner products are scaled before the $Softmax$ by the $\tau$ which is a learnable parameter that allows for a more precise or consistent distribution of attention weights. The new $A_{XC}\left(K,Q\right)$ operates along the dimensions of input vector d, which denoted the spectral information of the image, rather than along the amount of the patches n. Each output embedding is a convex combination of its corresponding embedding in V's d features. The computational cost is $O(d^2n)$ which has a linear computational computing complexity of input size.
Then, residual connection is used around each module followed by Layer Normalization to generate a deeper model \cite{Ba2016Layer}. For instance, each encoder block ($H^\prime$) can be written as:

\begin{equation}
H^\prime=\mathrm{\ LayerNorm\ }\left(\mathrm{\ SBA}\left(x\right)+x\right)
\end{equation}

\par Spatial Interaction block (SIB)
\par  Since SBA only focuses on spectral information not the spatial information between patches, a Special Interaction Block (SIB) is therefore introduced to enable explicit communication between patches. The SIB is built with two depth-wise $3\times3$ convolutional layers with Batch Normalization and ReLU non-linearity in between  \cite{Howard2017MobileNets:}. The output of the SIB can be written as:

\begin{equation}
\begin{gathered}
S=\operatorname{LayerNorm}\left(H^{\prime}+\right. \\
\left.\operatorname{Conv} 2\left(\text { BatchNorm }\left(\operatorname{ReLU}\left(\operatorname{Conv} 1\left(\operatorname{Conv} 1\left(H^{\prime}\right)\right)\right)\right)\right)\right)
\end{gathered}
\end{equation}

\paragraph{2) Multilayer Perceptron (MLP)}
\par A multilayer perceptron is a particular case of a feedforward neural network where every layer is a fully connected layer. As is common in transformer models, an MLP is added at the end of each encoder block, which contains two fully connected layers. 
While the SBA block restricts feature interaction within groups and the SIB can not allow for feature interaction, the MLP allows for interaction across all features. The output of MLP (F) can be written as:
\begin{equation}
F=\mathrm{LayerNorm\ }(S+Fc2(ReLU(Fc1(S)))
\end{equation}

\subsection{Local-to-Global Self-supervised learning}

\par In this work, to solve the data-hungry issue of deep learning models training, SSL is used to pretrain the proposed SSVT with unlabeled images.
Vision transformer is good at capturing long-range global spatial information. However, it fails to capture the local spatial information of small patches. To address this problem, we have proposed a local-to-global SSL method in which both local and global augmentations are performed to provide both global and local views of the input.  
The SSL consists of 2 main steps: the image augmentation and model training.

\par
\begin{figure}[h]
    \centering
    \includegraphics[width=0.4\textwidth]{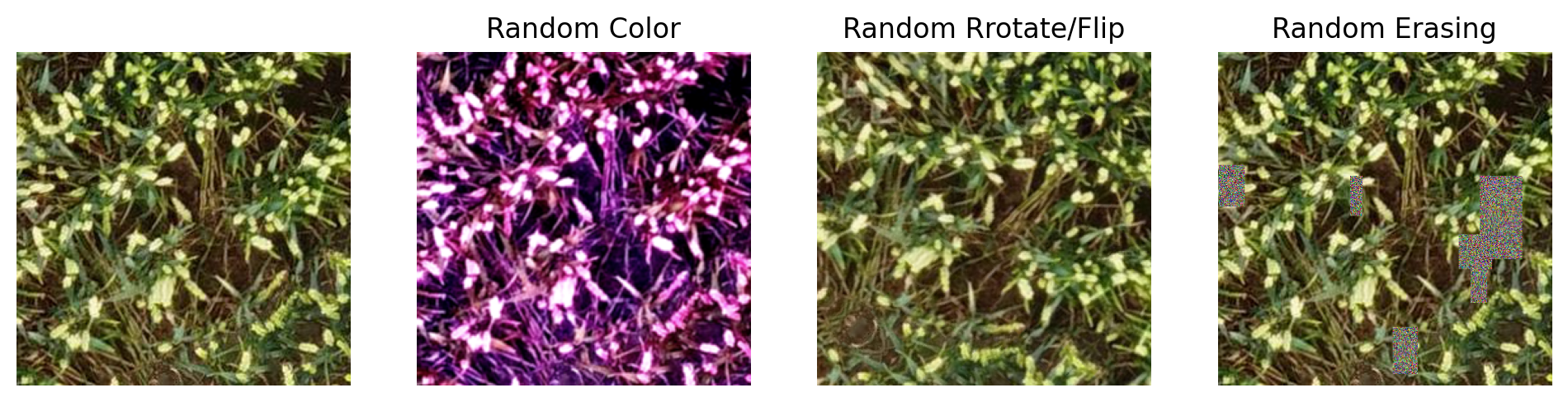}
    \caption{Image augmentations demonstration}
    \label{FIG:aug}
\end{figure}

 First, a random high-intensity image augmentation is used on the input images. The image augmentation technique is widely used for supervised and unsupervised training to improve the model's generalizability. It perturbs and modifies the data and keeps the output invariant, allowing the model to extract the most valuable features for classification. 
In this work, three types of image enhancement are first used for the input, including random color augmentation, random rotate/flip and random erasing. The random color augmentation consists of Brightness/ Contrast/ Saturation modifying, Color jittering, Gaussian blur, and Solarization. {Fig.~\ref{FIG:aug}} shows the three types of random image augmentation. This image augmentation strategy is also used for the supervised training.

\par
\begin{figure}[h]
    \centering
    \includegraphics[width=0.4\textwidth]{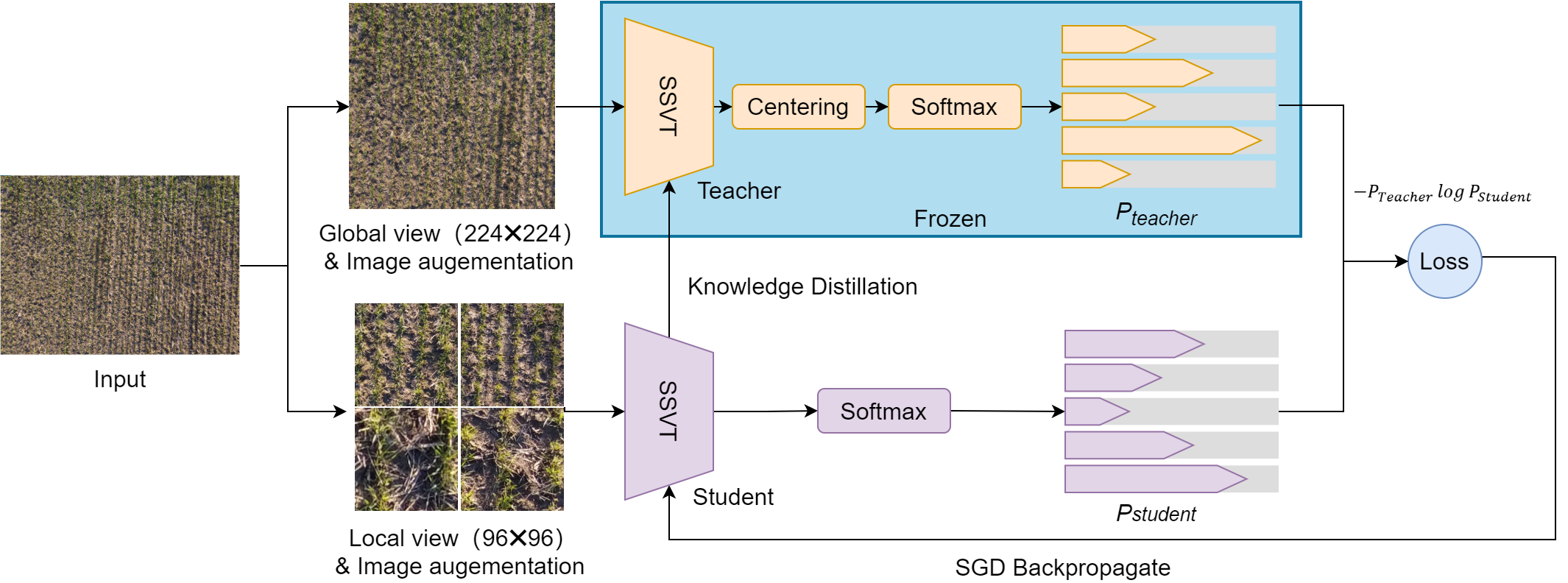}
    \caption{The flowchart of the Self-supervised learning}
    \label{FIG:4}
\end{figure}

 Then, we perform global and local augmentation on the same image to obtain both global and local views. The global views have an image size of $224 \times 224$. We assume that it contains the global context of the image. The small crops are called local views that have an image size of $96 \times 96$. It covers less than 50\% of the global view. We assume that it contains the local context. Then two views are fed into the SSL network. {Fig.~\ref{FIG:4}} shows the SSL framework. All local views are passed through the student network, while the global views are passed through the teacher network. It encourages the student network to interpolate context from a small crop and the teacher network to interpolate context from a bigger image. 

 \par The SSL network learns through a process called 'self-distillation' proposed by the paper ‘Be Your Own Teacher' \cite{Zhang2019Be}. There is a teacher and student network both using the proposed model SSVT. They have the same configuration with the same parameters and weights. The teacher is a momentum teacher, that all the weights are frozen and updated by students' weights ($\theta_s$) through an exponentially moving average. The update rule for the teacher's weights ($\theta_t$) is:

\begin{equation}
\theta_t\gets\lambda\theta_t+\left(1-\lambda\right)\theta_s
\end{equation}

\begin{figure}[h]
    \centering
    \includegraphics[width=0.4\textwidth]{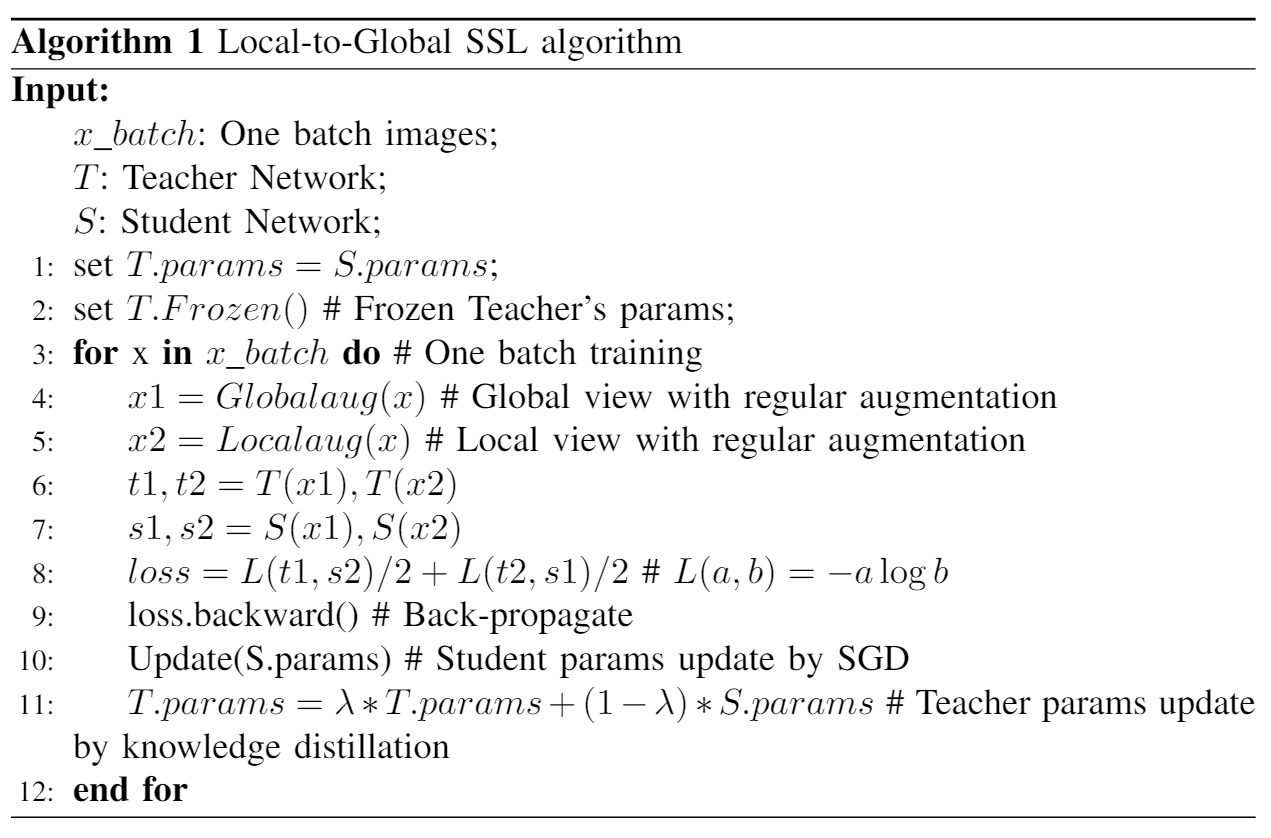}
    \caption{Local-to-Global SSL algorithm}
    \label{FIG:ssl}
\end{figure}

With $\lambda$ following a cosine schedule from 0.996 to 1 during training. The cross-entropy loss is used to make the two distributions the same, just as in knowledge distillation.

\begin{equation}
Loss=-P_{Teacher}log{P_{Student}}
\end{equation}
Centering \cite{Caron2021Unsupervised} is used to prevent the model from predicting a uniform distribution along all dimensions or dominated by one dimension regardless of the entry. The teacher’s raw activations ($A_t(x)$) have their exponentially moving average (c) subtracted from them. The center c is updated with an exponential moving average. The algorithm is shown in {Fig.~\ref{FIG:ssl}}. 
\begin{equation}
 A_t(x)\gets A_t(x)\ -c   
\end{equation}

\subsection{Model Evaluation}

\subsubsection{Experimental Design}

To evaluate the performance of the proposed SSVT, we have conducted three types of experiments: 1) Performance evaluation of SSVT for automated crop N prediction. 2) Ablation Study. 3) Evaluation of the generalizability of the proposed model using independent datasets.

\paragraph{1) Performance evaluation of SSVT for automated crop N prediction}
\par To evaluate the performance of the proposed SSVT for Crop N prediction, we first train the model based on the configuration (Section 3.4.3) with two different input sizes. The performance of the model including precision, recall and F1-score (Section 3.4.2) for each class and overall accuracy are reported. Then we compare the proposed SSVT with five state-of-the-art DL models. Two commonly used CNN based architectures, ResNet\cite{he2016deep} and EfficientNet\cite{tan2019efficientnet} with their state-of-the-art versions RegNet \cite{Radosavovic2020Designing} and EfficientNetV2, \cite{Tan2021EfficientNetV2:} along with ViT are selected for the performance comparison.  


\paragraph{2) Ablation Study}
\par In this case, two ablation studies are set to evaluate: 1) the performance of the proposed SSVT with and without SSL; 2) the impact of the spectral-spatial attention block.

\par The performance of the proposed SSVT with and without SSL
\par In this work, a local-to-global SSL method is proposed to pretrain the model on the unlabelled image generated from the drone. We evaluate the performance of the proposed SSVT trained from SSL and trained from scratch to show the impact of the SSL on model generalization. 

\par The impact of the spectral-spatial attention block
\par This work proposes the spectral-spatial attention block to replace the self-attention in the original ViT, making the attention module attend over the spectral and spatial information. In this case, we evaluate the effect of the proposed model, compared to the original vision transformer (The ViT-small with a similar number of parameters is selected in this work). 

\paragraph{3) Evaluation of the generalizability of the proposed model using independent drone datasets}

\par In this case, to evaluate the generalizability of the proposed SSVT model, we have evaluated the trained model on independent datasets. The images are captured from the drone in every growing stage, including Tillering \& Stem Extension, Heading \& Flowering and Ripening \& Maturity. 


\subsubsection{Evaluation metrics}
\par Accuracy, Precision, Recall, F1 score and the Confusion matrix are selected for the accuracy assessment to evaluate model performance. Accuracy is the most intuitive performance measure, and it is simply a ratio of correctly predicted observations to the total observations. The Precision measures the fraction of true positive detections, and the Recall measures the fraction of correctly identified positives. The F1-score considers both the Precision and the Recall to compute the score. The study establishes the classification matrices which Precision, Recall and F1-score calculated with the following equations:
\begin{equation}
    \mathrm{Accuracy\ }=\frac{tp+tn}{tp+tn+fp+fn}
\end{equation}

\begin{equation}
    Precision=\frac{tp}{tp+fp}
\end{equation}

\begin{equation}
    Recall=\frac{tp}{tp+fn}
\end{equation}

\begin{equation}
    {F1}_{score}=\frac{Recall\times P r e c i s i o n}{Recall+Precision}\times2
\end{equation}

Where True Positives (TP) are the correctly predicted positive values. True Negatives (TN) are the correctly predicted negative values. False positives and false negatives occur when the actual class contradicts the predicted class. False Positives (FP) means the predicted class is yes when the actual class is no. False Negatives (FN) means predicted class in no when actual class is yes.

\subsubsection{Experimental configuration}
\par This work aims to develop a new method for accurately estimating N status in crops (i.e., wheat in this case) based on crop images at the canopy scale. There are four types of N treatments, including High, Medium, Low and Control, our task is to classify the images into these four categories automatically. {Fig.~\ref{FIG:5}} b) shows images collected from different plots with different treatments. We randomly cropped them into $224 \times 224$ patches for the drone images and generated the unlabeled images for SSL. In this work, 4,800,000 images are generated. The SSL training uses the AdamW optimizer (Loshchilov and Hutter, 2017) and a batch size of 64, distributed over 3 GPUs (GeForce RTX 2080 Ti). The learning rate is linearly ramped up during the first ten epochs as 1e-3. After this warmup, we decay the learning rate with a cosine schedule. The weight decay also follows a cosine schedule from 0.04 to 0.4.

\par The detailed configuration of the proposed SSVT and ViT is shown in {Table.~\ref{table:2}}. It has 12 encoder layers. The dimension of keys is 384. To achieve the best performance of the model, we have tested it with two input sizes. As remote sensing data, the larger input covers a larger area and more spatial features can be covered. One input size is $224 \times 224$, which is the default input size for most deep learning algorithms. The other is $384 \times 384$, which is 1.5 times of the default size.

\begin{figure}[h]
    \centering
    \includegraphics[width=0.4\textwidth]{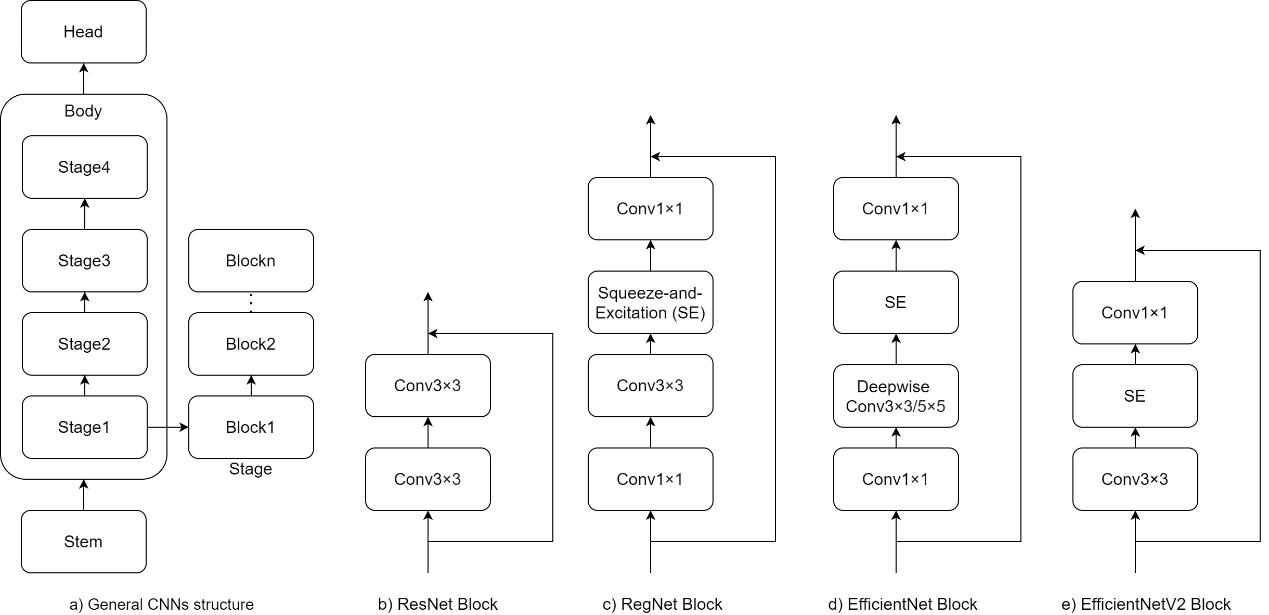}
    \caption{General network structure and detailed block structure for CNN models.}
    \label{FIG:7}
\end{figure}

\begin{table}[h]
\caption{The transformer model configuration}\label{table:2}
\centering
\resizebox{0.45\textwidth}{!}{%
\begin{tabular}{lllll}
\hline
\textbf{Model} & \textbf{Patch size} & \textbf{Layer} & \textbf{Dimension} & \textbf{Param (M)} \\ \hline
\textbf{ViT}   & 8                   & 12             & 384                & 21.67              \\
\textbf{SSVT}  & 8                   & 12             & 384                & 25.87              \\ \hline
\end{tabular}
}
\end{table}
The general network structure of selected models for comparison can be summarized as {Fig.~\ref{FIG:7}} a), which consists of a stem, followed by the body part, and head classifier (average pooling followed by a fully connected layer) that predicts output classes. The body part is composed of 4 stages that operate at progressively reduced resolution, and each stage consists of a sequence of identical blocks. The identical blocks of each model are shown in {Fig.~\ref{FIG:7}} b,c,d,e. For direct comparisons and to isolate benefits resulting from network design, the configuration of the models is based on the trained parameters, in which 20 million parameters are selected as the baseline in this work. Based on this, the ResNet with 50 layers, EfficientNet\_B5, RegNetY -4.0G, EfficientNetv2\_small are selected for comparison. 

\par For the supervised training on SSVT and the selected models for comparison. We first transfer the weights learned from SSL training to initialize the model. AdamW optimizer is used for 100 epochs using a cosine decay learning rate scheduler and 20 epochs of linear warm-up. A batch size of 64, a lower initial learning rate of 1e-4, and a weight decay of 0.05 are used for model training. The augmentation and regularization strategies used in this training to avoid the over-fitting, include, conventional image augmentation mentioned in section 3.2, random-size cropping, data mix-up\cite{zhang2017mixup} and label-smoothing \cite{muller2019does} regularization. We used five-fold cross-validation in this study. The dataset is divided into five groups at random, with four groups (80\% of dataset) utilised for training and the remaining groups used for testing each time. The average of the accuracies on the testing set over all folds is used to evaluate the classification performance.

\section{Result}

\subsection{Performance evaluation of SSVT for automated crop N prediction}

\par In this case, we report the performance of the proposed SSVT for automated crop N prediction with two input sizes ({Table.~\ref{table:3}}). With the input image size of $224 \times 224$, the accuracy of the proposed model reaches 0.962. With the input image size of $384 \times 384$, the accuracy of the proposed model reaches 0.965, which is slightly higher. For the rest of the evaluations and comparisons, we select $224 \times 224$ as the input size.

\begin{table}[h]
\caption{The proposed model performance comparison with different input sizes under four N treatments}\label{table:3}
\centering
\resizebox{0.45\textwidth}{!}{%
\begin{tabular}{lllll}
\hline
                                                   & \multicolumn{4}{c}{\textbf{Image size $224\times224$}} \\ \hline
\multicolumn{1}{c}{\textbf{Types of N treatments}} & Control     & Low       & Medium     & High     \\ \hline
\textbf{Precision}                                 & 0.949       & 0.992     & 0.985      & 0.923    \\
\textbf{Recall}                                    & 0.936       & 0.992     & 0.98       & 0.941    \\
\textbf{F1-score}                                  & 0.943       & 0.992     & 0.982      & 0.932    \\
\textbf{Accuracy}                                  & \multicolumn{4}{c}{0.962}                       \\ \hline
                                                   & \multicolumn{4}{c}{\textbf{Image size $384\times384$}} \\ \hline
\textbf{Precision}                                 & 0.973       & 0.991     & 0.982      & 0.917    \\
\textbf{Recall}                                    & 0.924       & 0.989     & 0.981      & 0.964    \\
\textbf{F1-score}                                  & 0.948       & 0.99      & 0.982      & 0.94     \\
\textbf{Accuracy}                                  & \multicolumn{4}{c}{0.965}                       \\ \hline
\end{tabular}
}
\end{table}

\begin{table}[h]
\caption{The performance comparison of the proposed SSVT with the existing conventional models.}\label{table:4}
\centering
\resizebox{0.45\textwidth}{!}{%
\begin{tabular}{llll}
\hline
                      & Param (M) & GFLOPs (GMac) & Accuracy (\%) \\ \hline
ResNet\_50            & 23.52     & 4.12          & 0.945         \\
EfficientNet\_B5      & 28.35     & 2.4           & 0.95          \\
RegNetY-4.0G          & 20.6      & 4.1           & 0.951         \\
EfficientNetv2\_small & 20.18     & 2.87          & 0.949         \\
ViT                   & 21.67     & 4.24          & 0.944         \\
SSVT                  & 25.87     & 4.71          & 0.962         \\ \hline
\end{tabular}
}
\end{table}

\par Meanwhile, we have compared our proposed model with the five most widely used CNN models. The results are shown in {Table.~\ref{table:4}}. With the lowest flops and the most parameters, EfficientNet reaches an accuracy of 0.95. The proposed model has intermediate parameters and the highest flops. The performance outperforms other approaches.

\subsection{Ablation Study}

\subsubsection{The performance of the proposed SSVT with and without SSL}

\par
\begin{figure}[h]
    \centering
    \includegraphics[width=0.4\textwidth]{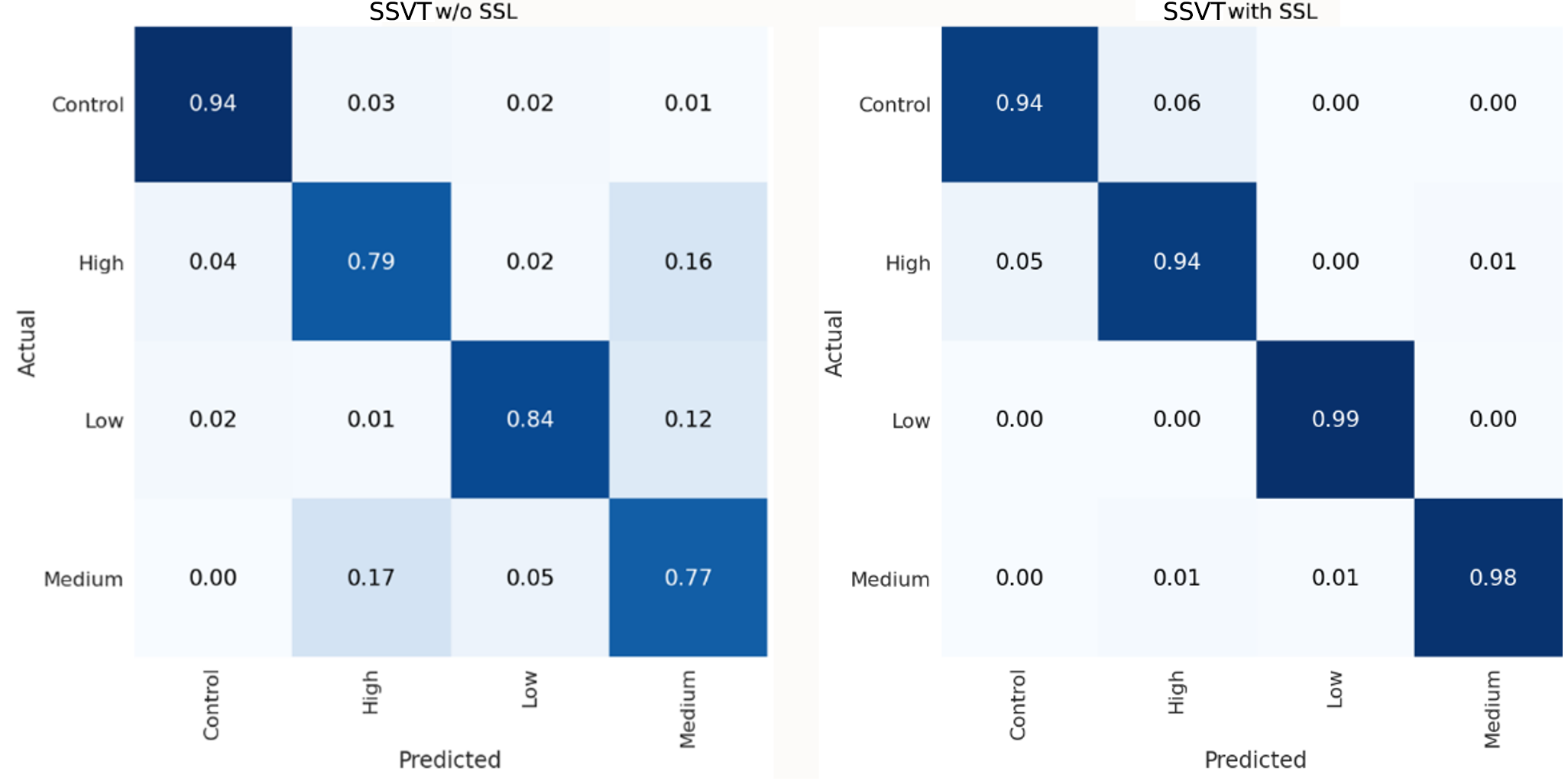}
    \caption{The Confusion matrix of the classification results achieved by the proposed SSVT (a) without and (b) with self-supervised learning}
    \label{FIG:9}
\end{figure}

 In this case, we train the model with initialized weights based on labelled data. The classification performance of the proposed model without SSL is reported in {Table.~\ref{table:5}} and {Fig.~\ref{FIG:9}}. Without the pretrained weights from SSL, the proposed SSVT cannot converge correctly. The Accuracy of the proposed model is only 0.836. As shown in {Fig.~\ref{FIG:9}}, the model without SSL performs well on N status (Control). However, it performs unsatisfactorily on other statuses including High, Low and Medium. Conversely, the model trained with SSL weights performs well on all N statuses.


\begin{table}[h]
\caption{Model performance without self-supersized learning for the four N treatments}\label{table:5}
\centering
\resizebox{0.45\textwidth}{!}{%
\begin{tabular}{lllll}
\hline
\textbf{Model}                                     & \multicolumn{4}{c}{\textbf{SSVT w/o SSL}} \\ \hline
\multicolumn{1}{c}{\textbf{Types of N treatments}} & Control   & Low     & Medium   & High    \\ \hline
\textbf{Precision}                                 & 0.937     & 0.898   & 0.738    & 0.784   \\
\textbf{Recall}                                    & 0.939     & 0.844   & 0.774    & 0.789   \\
\textbf{F1-score}                                  & 0.938     & 0.87    & 0.756    & 0.787   \\
\textbf{Accuracy}                                  & \multicolumn{4}{c}{0.836}                \\ \hline
\end{tabular}
}
\end{table}

\subsubsection{The impact of the spectral-spatial attention block}

\par In this case, to evaluate the effect of the proposed spectral-spatial attention block on vision transformer, we report the model performance of the proposed model and the original ViT. The results are shown in {Table.~\ref{table:6}}. It has demonstrated that our proposed model with spectral-spatial attention block has better classification performance for crop N status estimation than that of ViT.

\begin{table}[h]
\caption{Model performance comparison with the proposed model with SBA and the original ViT ( four types of N treatment: Control, Low, Medium and High)}\label{table:6}
\centering
\resizebox{0.45\textwidth}{!}{%
\begin{tabular}{lllll}
\hline
\textbf{Model}                                     & \multicolumn{4}{c}{\textbf{ViT}}  \\ \hline
\multicolumn{1}{c}{\textbf{Types of N treatments}} & Control  & Low   & Medium & High  \\ \hline
\textbf{Precision}                                 & 0.925    & 0.956 & 0.931  & 0.969 \\
\textbf{Recall}                                    & 0.994    & 0.975 & 0.944  & 0.865 \\
\textbf{F1-score}                                  & 0.959    & 0.965 & 0.938  & 0.914 \\
\textbf{Accuracy}                                  & \multicolumn{4}{c}{0.944}         \\ \hline
\multicolumn{1}{c}{\textbf{Model}}                 & \multicolumn{4}{c}{\textbf{SSVT}} \\
\textbf{Precision}                                 & 0.949    & 0.992 & 0.985  & 0.923 \\
\textbf{Recall}                                    & 0.936    & 0.992 & 0.98   & 0.941 \\
\textbf{F1-score}                                  & 0.943    & 0.992 & 0.982  & 0.932 \\
\textbf{Accuracy}                                  & \multicolumn{4}{c}{0.962}         \\ \hline
\end{tabular}
}
\end{table}

\subsection{Evaluation of the generalizability of the proposed model using independent datasets}

\par
\begin{figure}[h]
    \centering
    \includegraphics[width=0.4\textwidth]{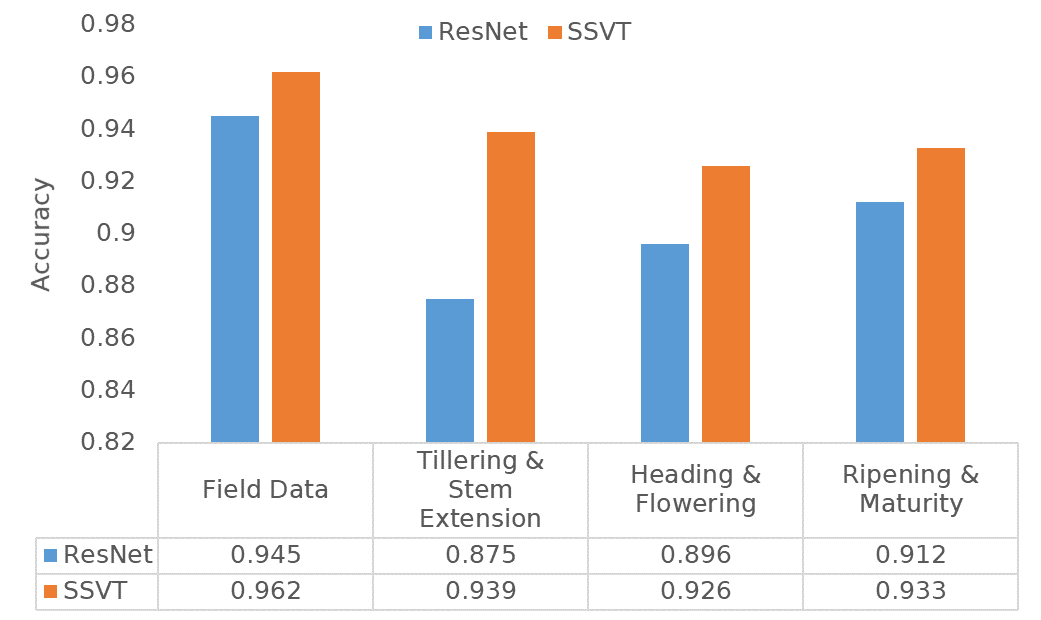}
    \caption{The model performance on independent drone datasets captured throughout all growing stages}
    \label{FIG:11}
\end{figure}

 To evaluate the generalizability of the proposed SSVT model, we evaluate the trained model on independent drone datasets captured at every growing stage, including Tillering \& Stem Extension, Heading \& Flowering and Ripening \& Maturity. The performance of the trained model on each growing stage are reported in {Fig.~\ref{FIG:11}}. 

\begin{figure}[htbp]
    \centering
    \includegraphics[width=0.4\textwidth]{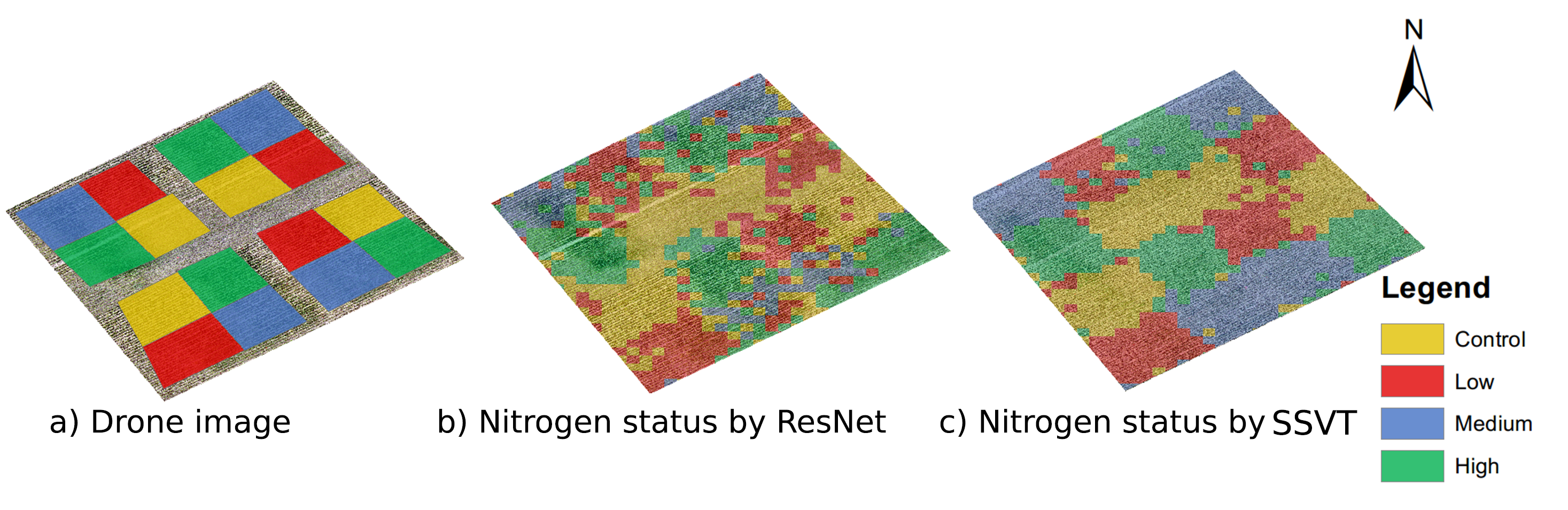}
    \caption{The nitrogen status estimation result on drone images captured at the tillering \& stem extension stage}
    \label{FIG:12}
\end{figure}

Compared to the model's accuracy based on the ground field data, the accuracy of proposed model on independent drone data is slightly decreased. The accuracy in Tillering \& Stem Extension, Heading \& Flowering, and Ripening \& Maturity is 0.939, 0.926 and 0.933, respectively. However, the performance of ResNet on independent drones dropped significantly. Compared to its accuracy on ground field data, it dropped by 0.07, 0.049 and 0.033 in all growing stages. {Fig.~\ref{FIG:12}} shows the N status estimation result on drone images captured at the early growing stage (Tillering \& Stem Extension). In the early stages of crop growth, all characteristics are not prominent. The estimated result of ResNet shows many misclassifications. The result of our proposed model is more precise and more accurate. The result shows the good generalizability of our proposed model.

\section{Discussion}
In this work, we propose a new deep learning-based method for accurately estimating the nitrogen status of wheat using images from UAV imagery named SSVT. Three experiments are designed to evaluate the performance of the model. Our discussions are based around the experiment results.

\subsection{Performance of SSVT for automated crop N prediction}

In the first experiment, we have evaluated the accuracy of the model and compared it with the existing state-of-the-art deep learning models including ResNet and its latest version RegNet, EfficientNet v1 and v2. The results demonstrate that our proposed model achieves an overall accuracy of 0.962. Under the similar parameters (20-30 millions), the classification accuracy of the proposed method outperforms the comparative structures. In general, the performance of model estimation on the nitrogen status of crops using RGB digital images might be affected by several factors, such as inconsistent image brightness and white balance in multiple observations, the shadow of crops and the soil background \cite{putra2020improving}. To avoid the effect of shadows on images, all the data in this work are taken at 11-12 AM to reduce the shadow of the plants and ensure sufficient light conditions. Moreover, we have performed high-intensity image color augmentation for model training, including Brightness/Contrast/Saturation, modifying, Color jittering and Solarization ({Fig.~\ref{FIG:aug}}). This is considered effective in improving the generalisability and robustness of the model \cite{shorten2019survey}, thus allowing it to maintain performance under different lighting conditions. 

\par
\begin{figure}[htbp]
    \centering
    \includegraphics[width=0.4\textwidth]{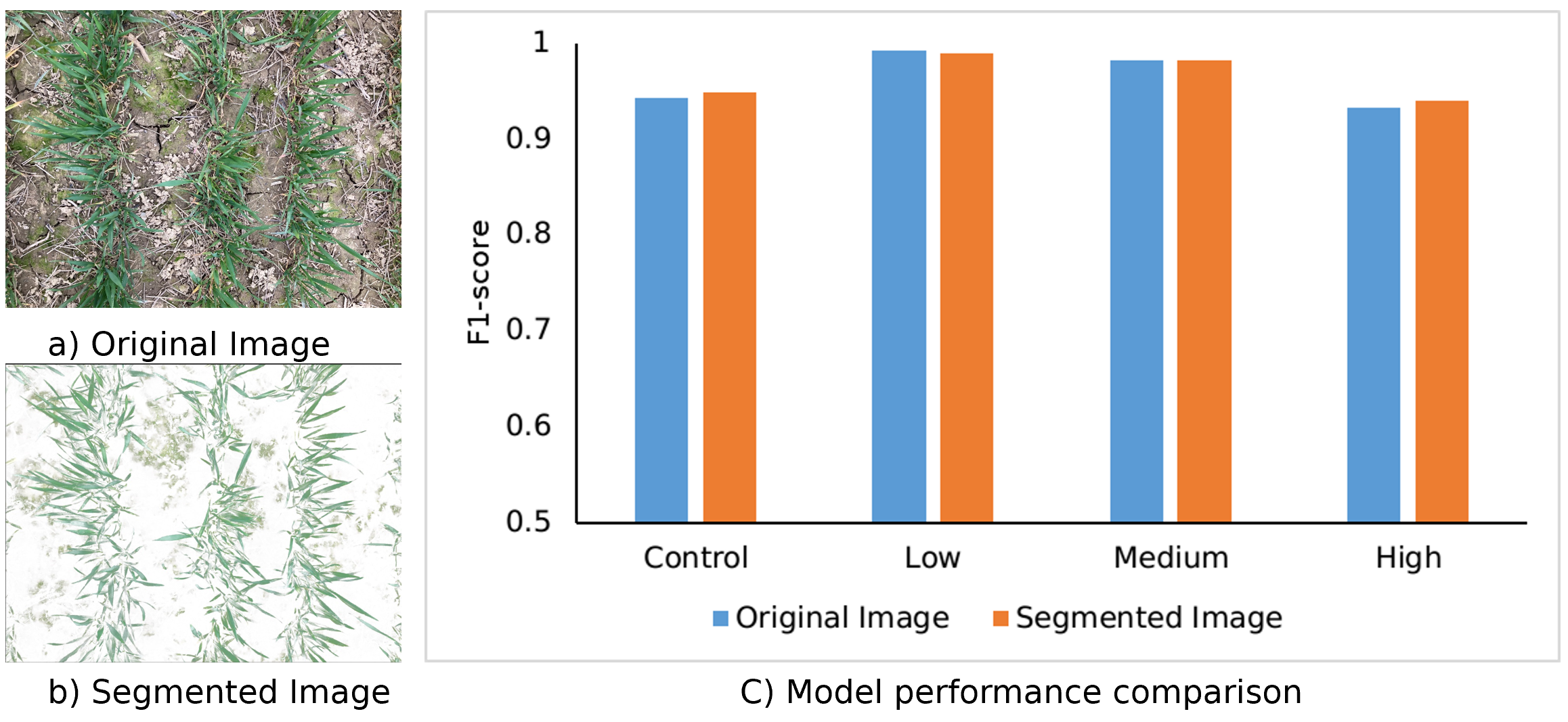}
    \caption{a) is the original image and b) is the segmented image without soil. c) shows the model performance with original and segmented image. }
    \label{FIG:cam}
\end{figure}

 For the impact of soil background, the typical method to remove the effect is to segment the soil from the image, as shown in {Fig.~\ref{FIG:cam}} a) and b). However, automatically identifying and segmenting crops from soil correctly in the high-resolution image is one of the most challenging problems in precision agriculture\cite{hernandez2016optimal,dyson2019deep}. In this work, a simple comparison experiment is performed to evaluate the performance using original and segmented images. The crop is segmented by a dynamic colour threshold through manual visual interpretation on each image. The F1-scores for each class of the models' performance are shown in {Fig.~\ref{FIG:cam}} c). The model using the segmented images does not improve the model's performance, and the results demonstrate that the proposed model has the ability to remove the effects of background from complex images.

\par
\begin{figure}[htbp]
    \centering
    \includegraphics[width=0.4\textwidth]{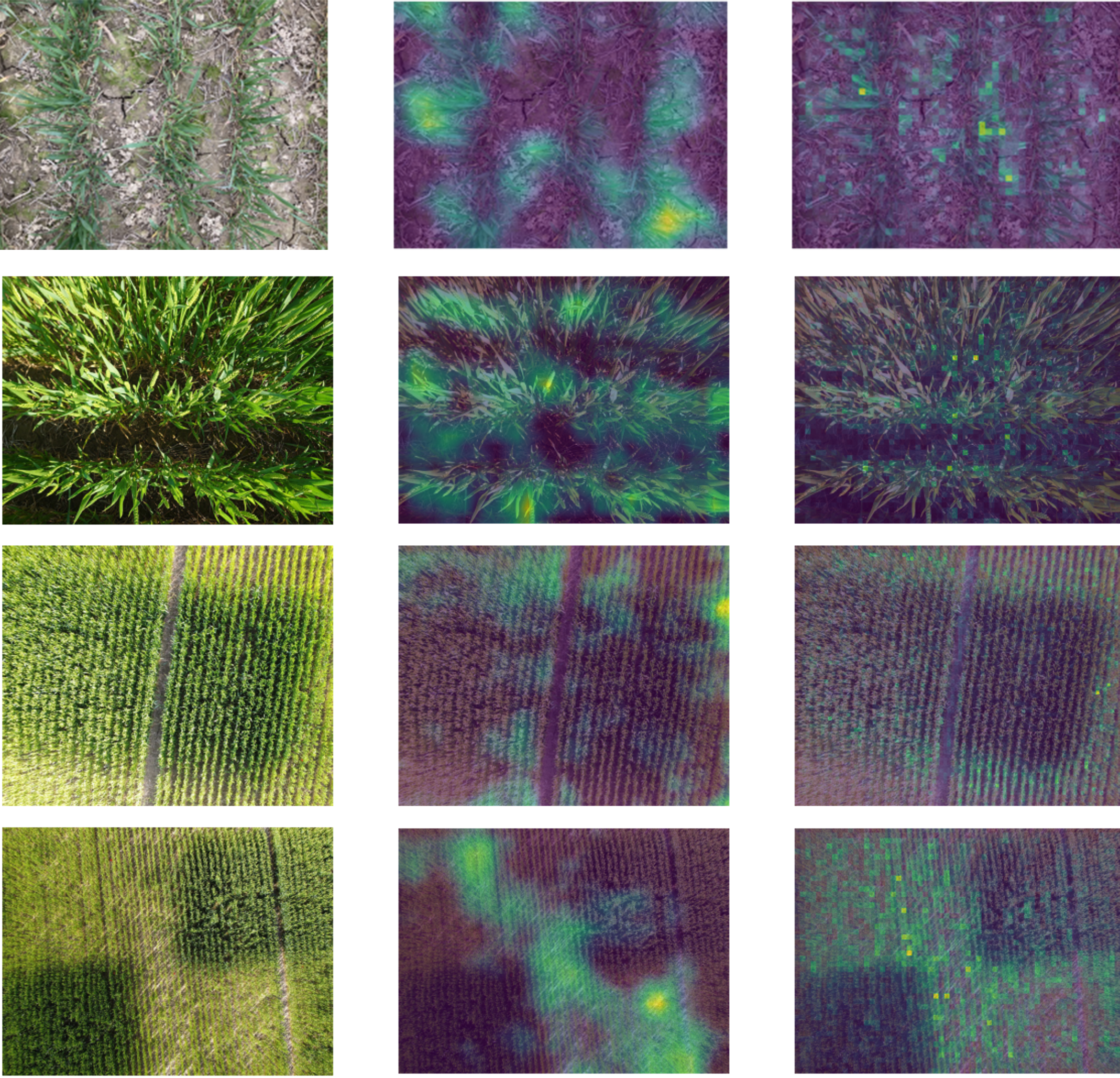}
    \caption{Visualization of the attention map for ResNet and proposed SSVT. The middle column represents the attention map of ResNet. The right column represents the attention map of SSVT}
    \label{FIG:13}
\end{figure}

 To deeply investigate the reasons for the ability to remove the background of our proposed SSVT, we visualize the attention map from the last block of the trained model to explain the decision-making area. The attention maps are shown in {Fig.~\ref{FIG:13}}. The middle column is the attention map of the ResNet model, and the right column is the map of the proposed SSVT. The images in the first two lines are captured in the field, and the attention maps show that the two models highlight the plant area from the first two lines. This result can explain why the soil background does not affect the deep learning-based model's performance on crop N status estimation.
The last two lines show the attention area on UAV images. The attention map of the proposed SSVT distinguishes well between regions with different N statuses. The attention map of ResNet also distinguishes between different regions, but not as clearly as the proposed SSVT. This leads to the higher accuracy of our proposed SSVT on independent drone data. Our third experiment evaluates the generalizability of the proposed model using independent UAV datasets. The results demonstrate that our proposed model outperforms the existing models in every growing stage.  

\subsection{Prospects and future work}

In this work, there are three main innovations of the proposed framework including 1) The SSL for model training with unlabeled datasets; 2) A novel spectral-spatial attention-based vision transformer network; 3) The computational complexity optimization on transformer network. 

\begin{figure}[h]
    \centering
    \includegraphics[width=0.4\textwidth]{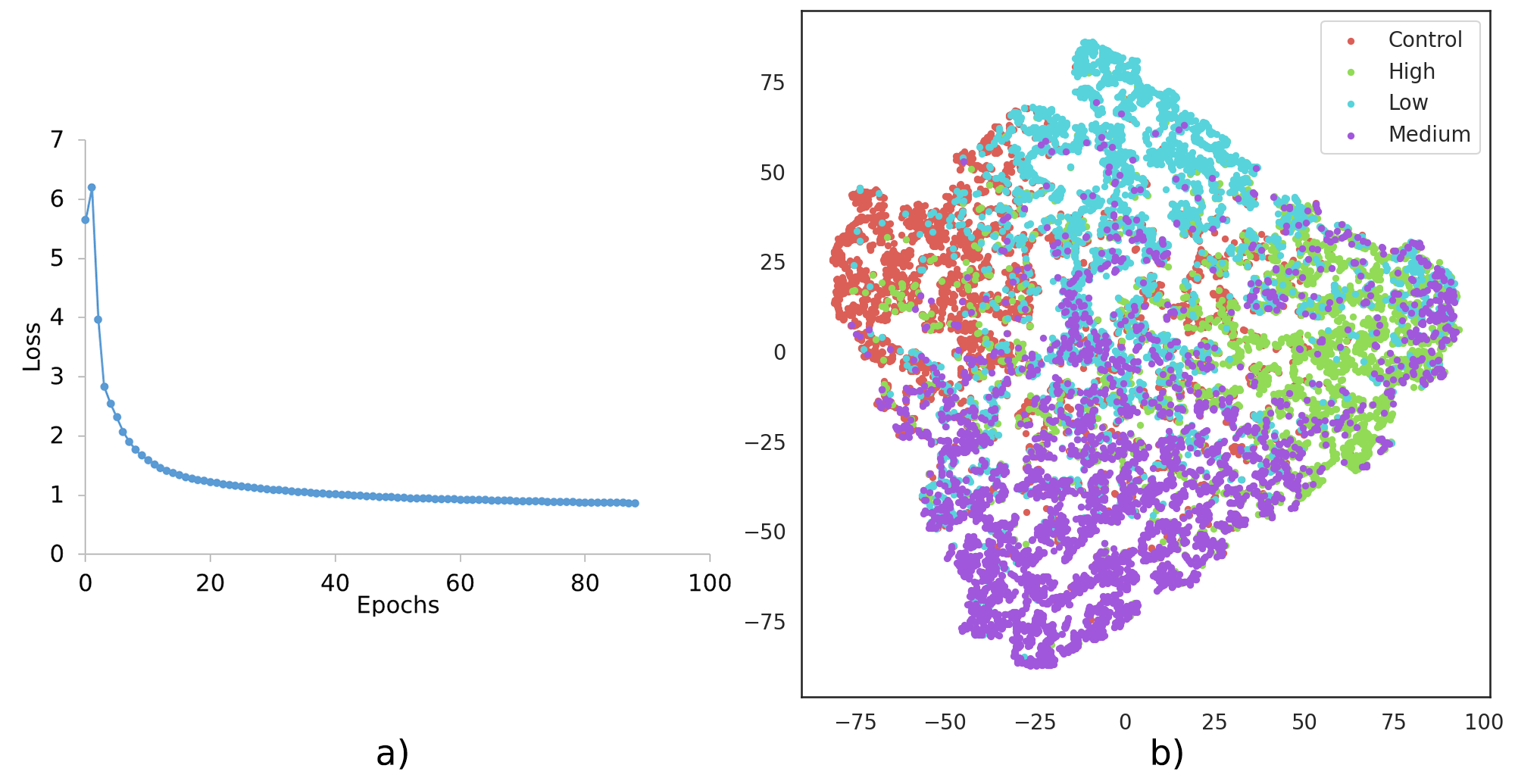}
    \caption{a) is the loss converge trend on self-supervised learning, b) is t-SNE visualization of the model trained with the proposed SSL.}
    \label{FIG:8}
\end{figure}

The first one is the SSL for model training with unlabeled datasets. In the second experiment, we evaluate the impact of the SSL. The result shows that the proposed model does not converge well when we train the model from init weight. The accuracy is only 0.836. However, when we trained the model with the weights from SSL, the model converged well and achieved the best performance. {Fig.~\ref{FIG:8}} a) shows the loss trend in SSL, which shows the model converges correctly. It indicates that the model can extract the similar features from the different views (‘local’ and ‘global’ views) of the same image. In {Fig.~\ref{FIG:8}} b), we visualize the features extracted from the labeled data based on the model trained with SSL by the t-SNE (van der Maaten Hinton, 2008). Although there are points that remain integrated with other points belonging to other classes, various clusters are easily recognized by different N statuses. This result explains why SSL can help the model train without labeled data. The proposed SSL is generic which can be applied to other applications with limited labeled datasets. Particularly, remote sensing applications often have large amounts of data but lack annotation. We believe our approach will significantly contribute to the remote sensing field through SSL from unlabeled data. 

\par The second innovation is that the proposed SSVT is capable of simultaneously capturing both spatial and spectral based features for accurate nitrogen diagnosis. In the experiment, we evaluate the performance of the proposed method compared to the existing vision transformer networks, focusing on spatial features only. The results indicate that by adding the SBA and SIB we proposed, our model achieves better accuracy. The SSVT is also a generic network. In this paper we have used it on RGB datasets and achieved satisfactory results. We believe it can be applied to multi to hyper spectral datasets and we will deliver this work in the future.  
\begin{figure}[htbp]
    \centering
    \includegraphics[width=0.4\textwidth]{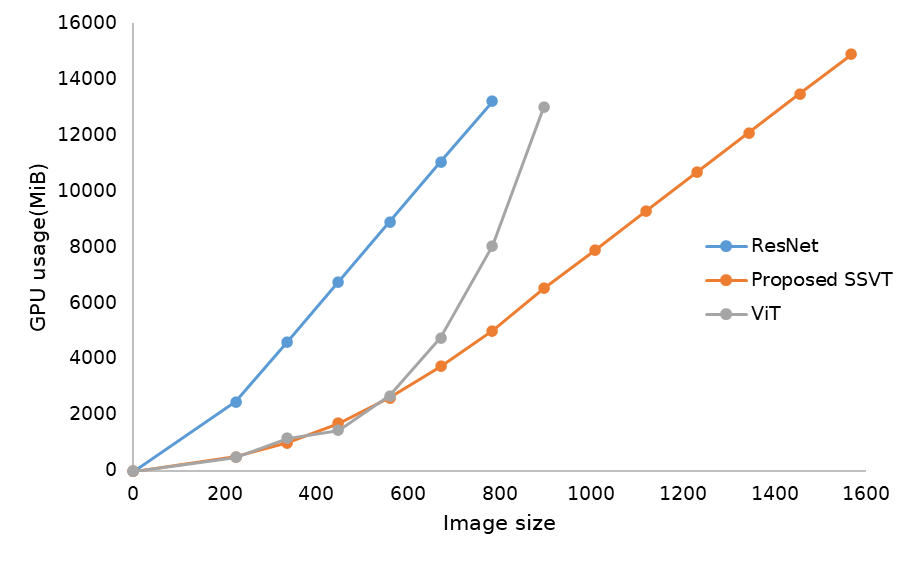}
    \caption{Inference GPU memory usage of ResNet, original VIT, and proposed SSVT}
    \label{FIG:10}
\end{figure}
\par The third innovation is the computational complexity optimization. The deep learning-based methods typically have millions of parameters, resulting in massive computational consumption. The original ViT has quadratic computational complexity to the image size due to the self-attention operation, which limits its usage on large images. In this work, the cross-covariance matrix is used to replace the gram matrix operation in the attention module. It changes the complexity of the transformer layer from quadratic to linear, which makes it possible for the model to handle large size images. We calculate and report the GPU usage of three models with increased input image size. We start from the commonly used size of $224\times224$ and then gradually and linearly increase the size of the input image (336, 448, 560 ……). The most widely used CNN model, Resnet with 50 layers and the original ViT, are selected for comparison. The inference GPU memory usage of the ResNet, the original VIT, and the proposed SSVT are shown in {Fig.~\ref{FIG:10}}. Our proposed SSVT has linear computational complexity with the size of the input image, which makes it possible to scale to a much larger image size (1600 x 1600 with 16GB GPU memory and $1344\times1344$ with 12GB GPU memory). The original VIT has quadratic computational complexity to the image size, which can only handle images with a size of $896\times896$ in 16 GB GPU memory and $784\times784$ in 12 GB GPU memory. Meanwhile, the proposed model has better computational efficiency and utilization than the CNN based model (ResNet).

\section{Conclusion}
We have proposed a novel spectral-spatial attention-based vision transformer (SSVT) for accurately estimating the nitrogen status of wheat using images from UAV imagery. The model framework proposes a spectral-spatial attention block consists of SBA and SIB, which can simultaneously learn both spatial and spectral features for accurate crop N estimation. The proposed model has been compared with state-of-the-art methods as well as being evaluated on both testing and independent datasets. The experimental results show competitive advantages over the existing works in terms of accuracy and computing performance, and model generalizability. Moreover, since model training requires massive labeled data, which is time consuming and costly. A local-to-global self-supervised learning has been introduced to pre-train the model with unlabeled data. We believe this approach will significantly contribute to the remote sensing field through self-supervised learning from unlabeled data. Meanwhile, the cross-covariance matrix is used to reduce the computational complexity of the model from quadratic to linear, which allows the proposed models to operate on a larger area. As a generic method, in the future, we will extend it to other data, especially multi to hyper spectral data to take advantage of its ability in both spectral and spatial feature learning.


%



\section*{Acknowledgment}
The work reported in this paper has formed part of the N2Vision project funded by UKRI-ISCF-TFP (Grant no. 134063).

\ifCLASSOPTIONcaptionsoff
  \newpage
\fi




%


\bibliographystyle{IEEEtran}
\bibliography{nitrogen}
%

\begin{IEEEbiography}[{\includegraphics[width=1in,height=1.25in,clip,keepaspectratio]{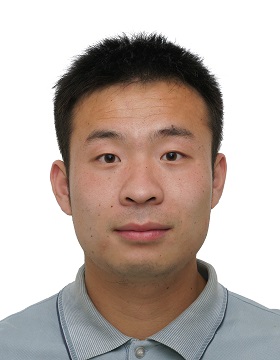}}]{Xin Zhang}
Xin Zhang is associate researcher in Manchester Metropolitan University (MMU), he received the B.S degree from The PLA Academy of Communication and Commanding, China, in 2009 and Ph.D. degree in Cartography and Geographic Information System from Beijing Normal University(BNU), China, in 2014. His current research interests include remote sensing image processing and deep learning.
\end{IEEEbiography}
\vskip -0.2in

\begin{IEEEbiography}[{\includegraphics[width=1in,height=1.25in,clip,keepaspectratio]{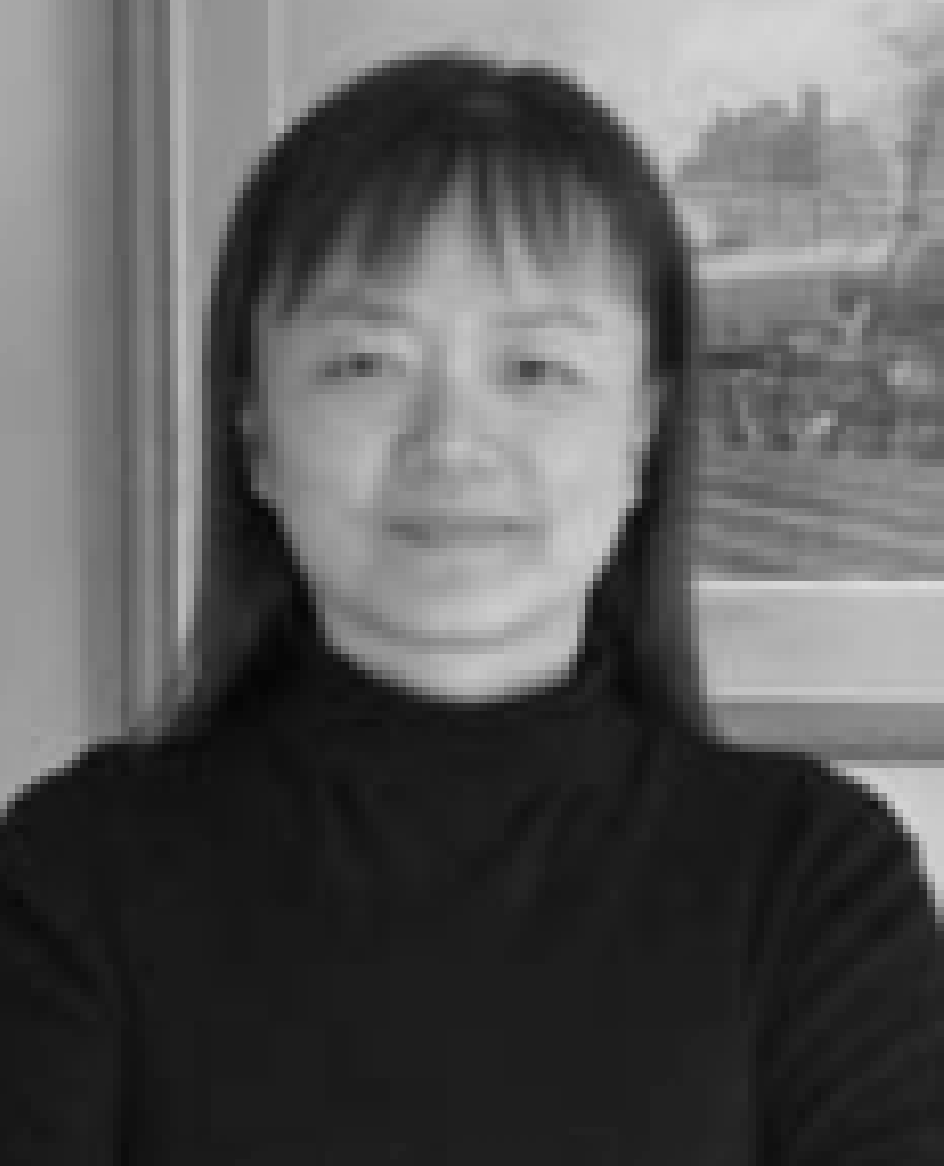}}]{Liangxiu~Han} received the Ph.D. degree in computer science from Fudan University, Shanghai, China, in 2002. She is currently a Professor of computer science with the School of Computing, Mathematics, and Digital Technology, Manchester Metropolitan University.  Her research areas mainly lie in the development of novel big data analytics and development of novel intelligent architectures that facilitates big data analytics (e.g., parallel and distributed computing, Cloud/Service-oriented computing/data intensive computing) as well as applications in different domains using various large datasets (biomedical images, environmental sensor, network traffic data, web documents, etc.). She is currently a Principal Investigator or Co-PI on a number of research projects in the research areas mentioned above.
\end{IEEEbiography}
\vskip -0.2in
\begin{IEEEbiography}[{\includegraphics[width=1in,height=1.25in,clip,keepaspectratio]{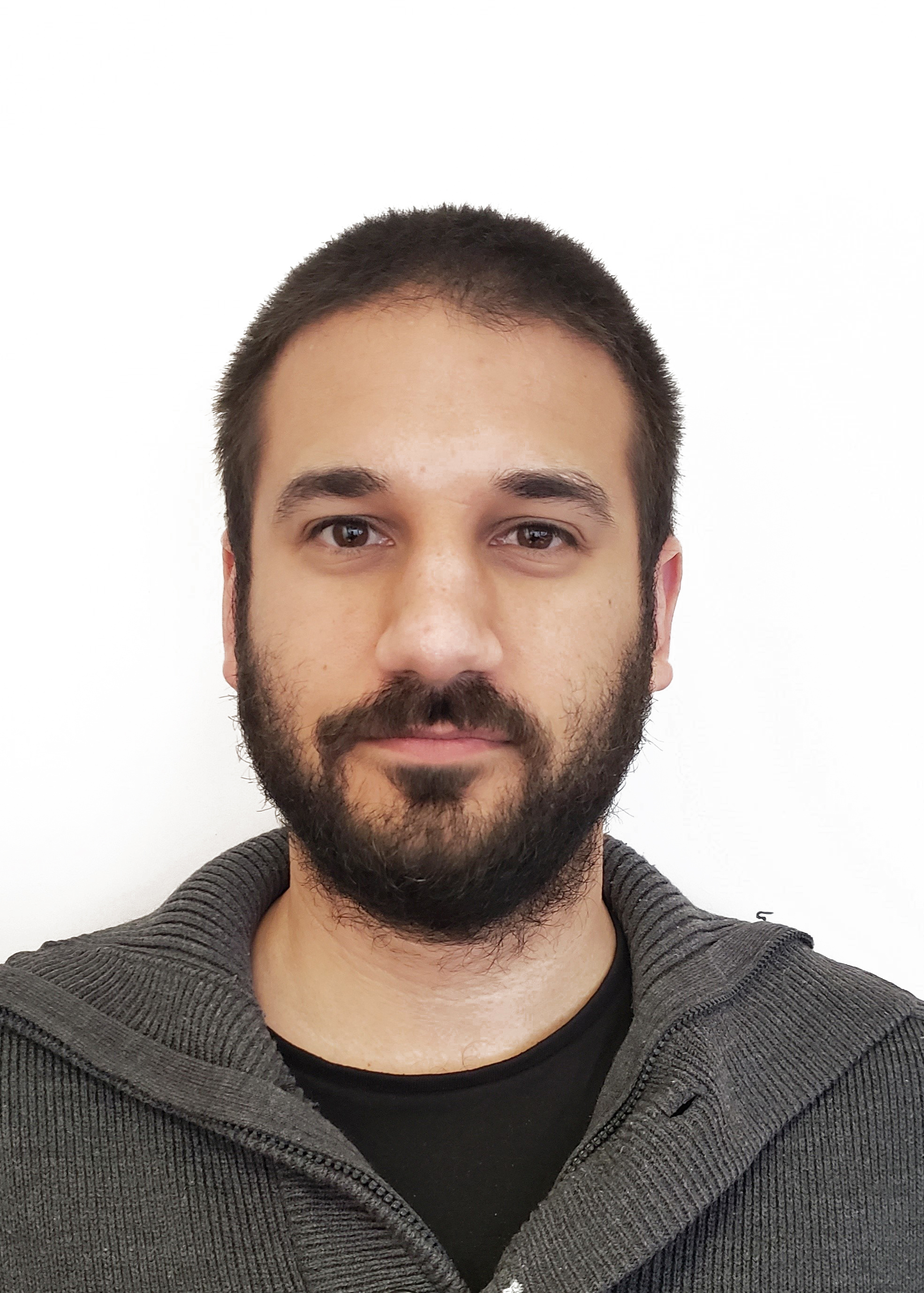}}]{Tam Sobeih}
Tam Sobeih is Research Associate with the School of Computing, Mathematics, and Digital Technology at Manchester Metropolitan University. His interests lie in providing real world solutions to complex problems, through the development of novel intelligent architectures, big data analytics and artificial intelligence applications, leveraging his prior experience in industry.
\end{IEEEbiography}
\vskip -0.2in

\begin{IEEEbiography}[{\includegraphics[width=1in,height=1.25in,clip,keepaspectratio]{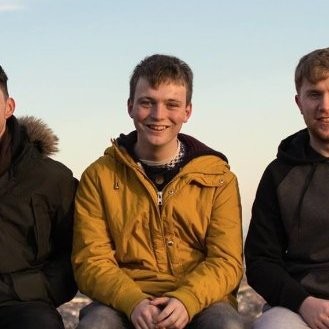}}]{Lewis Lappin}
Lewis is machine learning engineer at GMV. He graduated from University of Edinburgh in Computational Physics looking at Online Event Classification for trigger using Deep Learning at DUNE.

\end{IEEEbiography}
\vskip -0.2in
\begin{IEEEbiography}[{\includegraphics[width=1in,height=1.25in,clip,keepaspectratio]{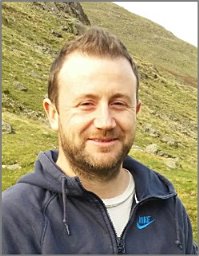}}]{Mark Lee}
Mark Lee is Research Fellow in Natural Capital and Plant Health at the Royal Botanic Gardens Kew. He use state of the art technologies and techniques, including plant growth chamber and field experiments, robotics, networked microsensors, machine learning and advanced data modelling to research livestock and arable farming. He is particularly interested in using these techniques to generate new ideas which can be used on farms to increase food production, reduce environmental degradation, tolerate our changing climate and increase the delivery of other ecosystem services.
\end{IEEEbiography}
\vskip -0.2in
\begin{IEEEbiography}[{\includegraphics[width=1in,height=1.25in,clip,keepaspectratio]{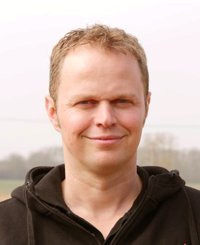}}]{Andew Howard}
Andew Howard is farmer. Consultant with Abacus Agriculture and Groundswell Agronomy. Nuffield Scholar. Into No Till, Soil Health, cover crops, intercropping and reduced inputs.
\end{IEEEbiography}
\vskip -0.2in
\begin{IEEEbiography}[{\includegraphics[width=1in,height=1.25in,clip,keepaspectratio]{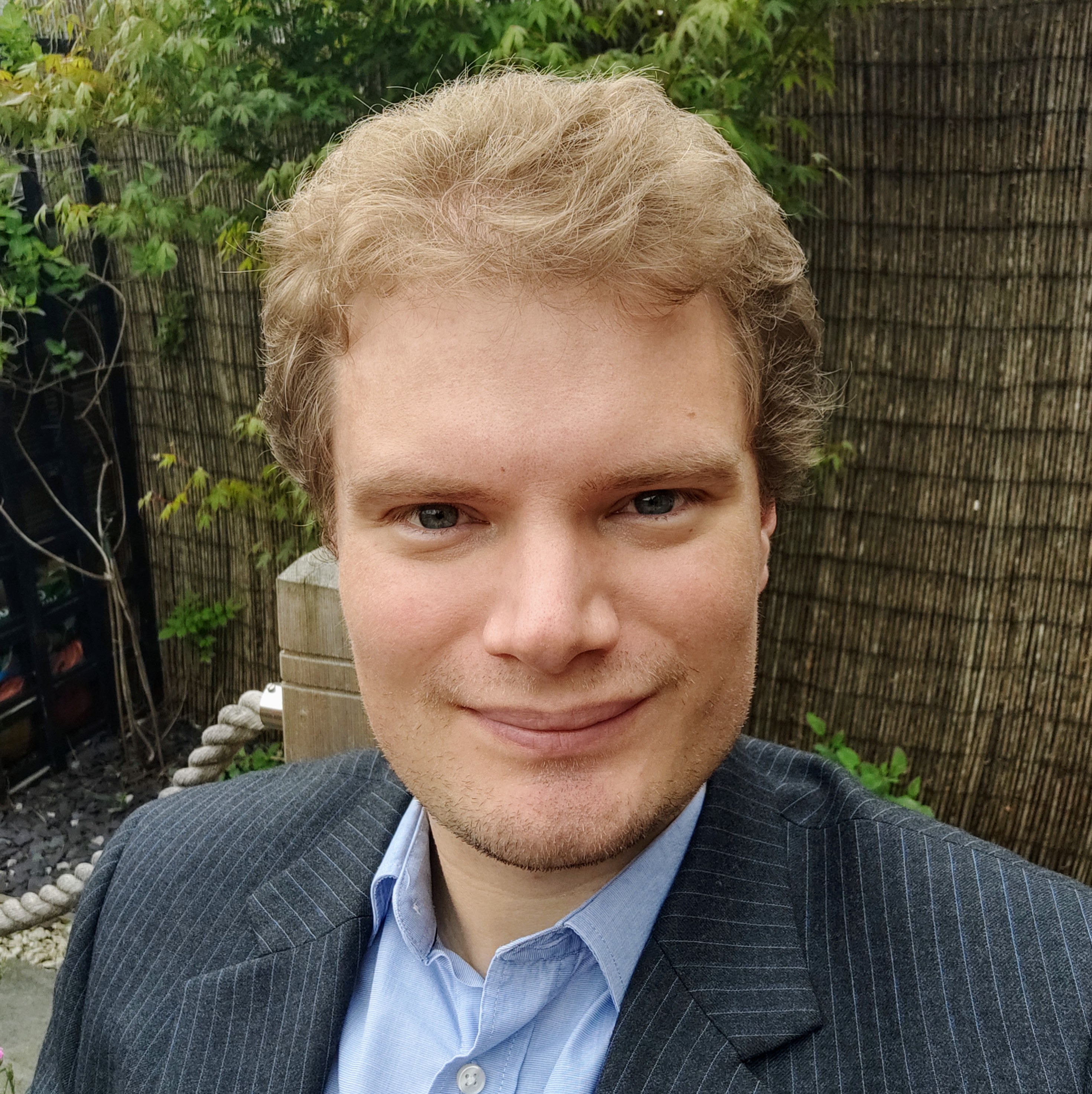}}]{Aron Kisdi}
Aron Kisdi is a chartered engineer by the Royal Aeronautical Society and holds a Masters degree in Space Systems Engineering from university of Southampton. With 12 years of experience in industrial research of robotics in particular autonomy and AI, Aron is currently working on developing robotics  and autonomous systems for space exploration as well as application of autonomy technology for use cases on Earth in particular agriculture and mining. He focuses of efficiency and robustness of these systems and believes commercial and R\&D success has to compliment one another.
\end{IEEEbiography}



\end{document}